\pdfoutput=1

\documentclass[11pt]{article}

\usepackage[]{acl}

\usepackage{times}
\usepackage{latexsym}
\usepackage{graphicx}
\usepackage{amssymb}
\usepackage{amsmath}
\usepackage{amsthm}
\usepackage{booktabs}
\usepackage{algorithm}  
\usepackage{algorithmic}
\usepackage{placeins}
\usepackage{multirow}
\usepackage{ulem}
\usepackage{subfigure}
\usepackage{enumitem}
\usepackage{bbding}
\usepackage{pifont}
\usepackage{wasysym}
\usepackage{caption}
\usepackage{xcolor} 
\usepackage{soul} 
\definecolor{revision}{RGB}{187,216,151} 
\definecolor{darkgreen}{rgb}{0.518, 0.604, 0.408}
\definecolor{darkpink}{rgb}{0.85, 0.490, 0.718}
\definecolor{mypurple}{rgb}{0.800,0.672,0.973}

\newtheorem{problem}{Problem}
\usepackage[T1]{fontenc}

\usepackage[utf8]{inputenc}

\usepackage{microtype}

\usepackage{inconsolata}

\usepackage{url}

\title{ClaimGen-CN: A Large-scale Chinese Dataset for Legal Claim Generation}

\author{Siying Zhou$^1$, Yiquan Wu$^1$, Hui Chen$^1$, Xavier Hu$^{1}$, Kun Kuang$^{1*}$\\
\textbf{Adam Jatowt$^{2}$, Ming Hu$^{1}$, Chunyan Zheng$^{1*}$, Fei Wu$^{1}$}\\
$^1$Zhejiang University, Hangzhou, China\\
$^2$University of Innsbruck, Innsbruck, Austria\\
\small\texttt{\{zhousiying, wuyiquan, 22402119, kunkuang, hm606, boxzheng\}@zju.edu.cn}\\
\small\texttt{xavier.hu.research@gmail.com, 
adam.jatowt@uibk.ac.at, wufei@cs.zju.edu.cn}}

\begin{document}
\maketitle
\begin{abstract}

Legal claims refer to the plaintiff's demands in a case and are essential to guiding judicial reasoning and case resolution. While many works have focused on improving the efficiency of legal professionals, the research on helping non-professionals (e.g., plaintiffs) remains unexplored.
This paper explores the problem of legal claim generation based on the given case's facts.
First, we construct ClaimGen-CN, the first dataset for \textbf{\uline{C}}hi\textbf{\uline{n}}ese legal \textbf{\uline{claim}} \textbf{\uline{gen}}eration task, from various real-world legal disputes.
Additionally, we design an evaluation metric tailored for assessing the generated claims, which encompasses two essential dimensions: factuality and clarity.
Building on this, we conduct a comprehensive zero-shot evaluation of state-of-the-art general and legal-domain large language models.
Our findings highlight the limitations of the current models in factual precision and expressive clarity, pointing to the need for more targeted development in this domain. ClaimGen-CN dataset is available at: \url{https://github.com/JosieZhou00/ClaimGen-CN}.

\end{abstract}

\renewcommand{\thefootnote}{\fnsymbol{footnote}}
\footnotetext[1]{Corresponding Authors.}
\renewcommand{\thefootnote}{\arabic{footnote}}

\section{Introduction}

Over the past decades, the advancement of natural language processing (NLP) techniques has advanced the field of Legal Artificial Intelligence (Legal AI). Legal AI is an important subfield of artificial intelligence, with the goal of supporting individuals across various legal tasks, including legal judgment prediction \citep{zhong2018legal, xu2020distinguish, wu2023precedent}, court view generation \citep{wu2020biased}, similar case matching \citep{bhattacharya2020hier}, legal language understanding \citep{chalkidis2021lexglue}, and legal question answering \citep{zhong2020jec}. As Legal AI systems become more advanced, it is essential to reflect not only on what these systems can do, but also on what they ought to contribute to society. If we could push the boundaries of what we envision AI to be, to the point where we explicitly require it to have a “positive impact on people and communities,” ensuring our definition of success explicitly includes that, AI could make the world a better place \citep{liWorldsSeeCuriosity2023}. Yet, the existing research predominantly targets courtroom trials and judge assistance \citep{ma2021legal,malik-etal-2021-ildc,feng2022legal,zhang-etal-2023-fedlegal,le2024topology,li-etal-2025-legal}, with limited attention to pre-trial contexts or non-professional needs, such as legal claim generation.

\begin{figure*}[t]
\centering
\includegraphics[scale = 0.39 ]{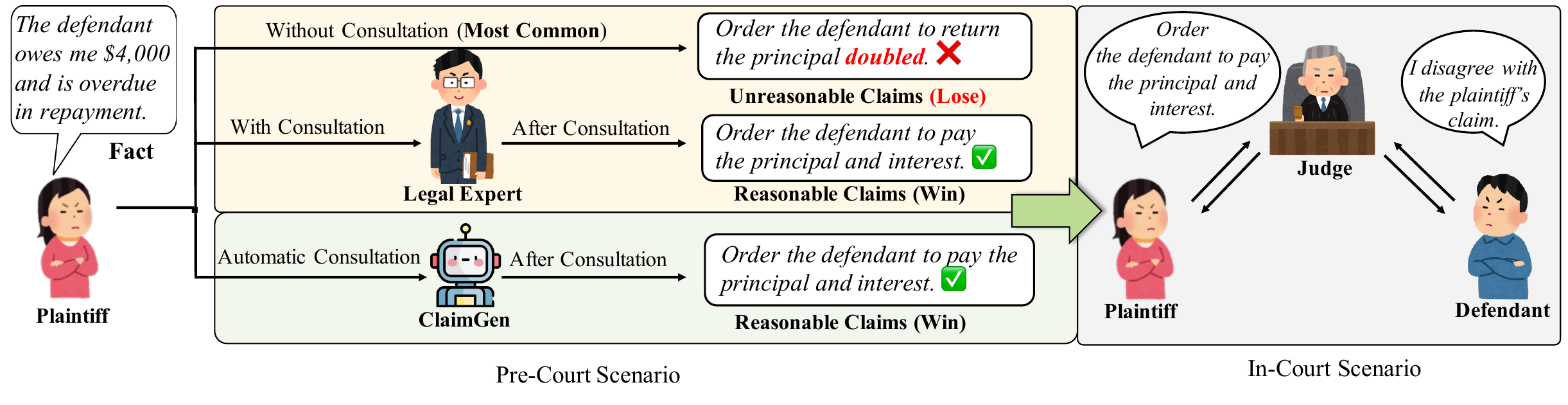}
\caption{
Conceptual overview of the differences between the pre-court scenario (left part) and the in-court scenario (right). In the pre-court scenario, claims are generated and prepared, which are then addressed and resolved during the in-court scenario. It is important to note that unreasonable claims may lead to a loss.}
\label{fig:example}
\vspace{-10pt}
\end{figure*}

To promote the rule of law and make legal support more accessible, we advocate exploring the problem of legal claim generation. As Figure \ref{fig:example} shows, different legal scenarios play different roles. The pre-court scenario is primarily dedicated to preparing claims for the plaintiffs, whereas the in-court scenario is where those claims are utilized. Our work is concentrated on the pre-court phase, focusing on the interests of the parties involved. To the best of our knowledge, we are the first to explore the problem of legal claim generation and to reaffirm its significance in civil litigation. 

Technically, legal claim generation introduces two key challenges that set it apart from prior Legal AI tasks. First, it is inherently open-ended: models must generate legal claims directly from factual narratives, without the guidance of predefined templates, as is common in judgment generation. Second, the input is typically authored by non-experts. Unlike court view generation, which builds on legally structured texts written by judges, this task requires interpreting informal, often unstructured, and emotional descriptions. The model must extract relevant facts, infer the underlying legal intent, and articulate it as a clear and valid claim.

To support research on legal AI in the field of civil litigation, several legal datasets have been constructed \citep{wang2018modeling, long2019automatic, xiao2021lawformer}. However, existing work still faces the following challenges: 1) Insufficient Data Coverage: Many existing datasets are limited to one or a few causes of action. For instance, both AC-NLG \citep{wu2020biased} and MSJudge \citep{ma2021legal} focus only on the private lending category in civil cases. 2) Lack of diverse and fine-grained metrics: Commonly used metrics like BLEU \citep{papineni2002bleu} and ROUGE \citep{lin-2004-rouge} focus mainly on n-gram overlap and are not comprehensive enough for evaluating the completeness of key information in legal texts.

To address the limited scope and diversity of existing legal datasets, we manually construct a dataset from real-world cases covering one hundred causes of action. Moreover, to better evaluate model performance in claim generation, we introduce several fine-grained criteria, which include two essential dimensions: factuality and clarity. To better understand current model capabilities, we further conduct a comprehensive zero-shot evaluation of state-of-the-art general and legal-domain large language models on the legal claim generation task.

To summarize, our main contributions are as follows:
\begin{itemize}

    \item We explore legal AI from the perspective of non-professionals (e.g., plaintiffs), and extend its scope to a novel and significant problem: legal claim generation. 
    
    \item We construct a large-scale Chinese dataset, ClaimGen-CN, built from raw civil legal documents. Furthermore, we introduce two key aspects for assessing claims: factuality and clarity.
    
    \item We conduct a comprehensive zero-shot evaluation of state-of-the-art general and legal-domain large language models on the legal claim generation task. Our detailed error analysis reveals notable limitations in current models, particularly in terms of factual accuracy and linguistic clarity.

\end{itemize}

\section{Related Work}

\subsection{Legal AI}
Legal AI aims to enhance tasks within the legal domain through the utilization of artificial intelligence techniques \citep{zhong2020does,katz2023natural}. The field has evolved from foundational developments in legal question answering and entity recognition, as explored by early researchers like \citet{monroy2009nlp} and \citet{cardellino2017legal}, to more advanced applications. The integration of AI in areas such as court view generation and legal summarization has shown its potential to process and interpret complex legal texts \citep{wu2020biased,hachey2006extractive}. Recent advancements in legal language understanding and predictive models for legal judgments demonstrate the significant strides made in applying AI to legal analysis \citep{bommarito2018lexnlp,zhong2018legal,liu2023ml}. The ongoing innovations in the field, particularly in the development of legal pre-trained models and event detection in legal documents, are further evidenced in the research by \citet{chalkidis2020legal}, showcasing the transformative impact of AI on legal research and practice.

\subsection{Legal Datasets}
Recent advancements in Legal AI have been significantly driven by the development and use of specialized legal datasets. For instance, the European Court of Human Rights (ECHR) dataset provided rich text data for human rights case law research, further diversifying the applications of AI in legal research \citep{aletras2016predicting}. LEXGLUE dataset provided a comprehensive benchmark for various legal NLP tasks, offering an invaluable resource for testing and improving AI models in the legal domain \citep{chalkidis2020legal}. Additionally, the Contract Understanding Atticus Dataset (CUAD) was designed for AI-based contract review and analysis, enabling more accurate and efficient processing of legal contracts \citep{hendrycks2021cuad}. 
TWLJP dataset was collected using indictments to assist prosecutors in indexing charges and crime factual descriptions \citep{chien2024legal}.

However, existing works primarily aim to assist experts such as judges and prosecutors, often overlooking common individuals in need of help with regards to their infringed rights. Our work shifts the research perspective from aiding legal professionals to assisting the non-expert public.

\section{Problem Formulation}

We first clarify the definition of the terms as follows.

$\bullet$ \textbf{Facts and Reasons} consist of two parts: a brief narrative detailing the dispute's background, including the legal relationship, time, place, specific case details, cause, process, circumstances, and consequences; and the grounds upon which the plaintiff seeks court relief. Evidence is encompassed in the plaintiff’s presentation of facts and reasons. We define facts and reasons as a token sequence $f$.

$\bullet$ \textbf{Claims} can be singular or multiple, each representing a specific demand to the court for the protection of civil rights. A claim is defined as a token sequence $c$.
For a given case, there can be several claims, which can be denoted as $C=\{c_1, c_2, ...,c_n\}$, where $n$ is the number of claims.

Common types of legal claims include: 1) Confirmation of Legal Relationships. This claim seeks court confirmation for certain legal relationships, such as the confirmation of the status of a missing or deceased individual.
2) Enforcement of Obligations. This claim compels the opposing party to fulfill payment obligations. Examples include seeking compensation for losses, demanding the repayment of loan principal and interest, or demanding compliance with contractual obligations.
3) Alteration or Termination of Civil Legal Relationships. This claim covers demands for changes or termination of specific civil legal relationships. Examples include filing for divorce or demanding cancellations of contracts.

Then the problem can be defined as:
\begin{problem} [Claim generation]
Given the facts and reasons $f$, the task is to predict the claims $C$.
\end{problem}

\section{Dataset}

\subsection{Dataset Construction}

We construct ClaimGen-CN, a Chinese dataset for legal claim generation, from 207,748 civil documents sourced from China Judgments Online\footnote{\url{http://wenshu.court.gov.cn/}}. While these documents span civil, criminal, and administrative cases, our work specifically focuses on civil cases for two main reasons.

$\bullet$ Civil cases, distinct from criminal or administrative cases, often involve private litigants and a wider variety of claims, making them more relevant for our research on claim generation. In criminal cases, since most claims are brought by professionals (prosecutors) and private prosecutions are rare \citep{krauss2009theory, Xiong2021Relationship}, the generated claims carry a lesser significance in promoting the accessibility of judicial assistance. In administrative litigation, the court's decisions are not strictly bound by the plaintiff's claims \citep{he2022}; unlike civil litigation, which follows the ``plaintiff requests, and the court adjudicates accordingly'' approach, the generated claims do not have as significant an impact on judges in administrative cases as they do in civil cases.

$\bullet$ Civil cases constitute the majority of the three categories of cases, accounting for 87\% of all cases, followed by criminal cases at 10\%, and administrative cases at 3\%, according to publicly available data\footnote{The proportions of the three case types are calculated based on statistics from the China Judgments Online website as of December 6, 2023: 87,618,891 civil documents, 10,010,356 criminal documents, and 3,031,224 administrative documents.}. Furthermore, civil litigation is the most common legal proceeding encountered in everyday life.

\subsubsection{Raw Data Processing}

Each initial document contains unstructured content over the entire document along with additional information, including the cause of action, title, start time, and end time. To maintain material homogeneity, we only select first-instance civil judgment documents and filter out those that are not publicly available for some reason. To ensure completeness, we segment the content, retaining only the task-related segments. To assess completeness, we use a sequential inclusion criterion with the following keywords: ``file a lawsuit with this court'', ``facts and reasons'', ``argument'', ``this court believes'', and ``judgment as follows''. The content has been then segmented into specific sections, including introduction, plaintiff's facts, plaintiff's claims, defendant's arguments, court's findings, and judgment. For our task, we exclusively utilize plaintiff's facts as input and plaintiff's claims as output.

\subsubsection{Structured Data}

Among all the sampled data, there are 134 causes of action. Due to insufficient data in some categories, we retain the top 100 most common civil causes of action for our main dataset, which we call \textbf{ClaimGen-CN}. In addition, we constructed a test set, \textbf{ClaimGen-CN-test}, by exclusively selecting cases where the court fully supports the plaintiff's claims.  This allows us to obtain reference claims that are both reasonable and legally grounded. This test set still follows the original distribution of causes of action from the sampled data.

Table \ref{tab:stat} presents the statistics of the processed datasets; all the experiments are conducted on the same datasets. 

\begin{table}[ht]
\centering
\small
\resizebox{\linewidth}{!}{
\begin{tabular}{lcc}
\toprule[2pt]
\textbf{Type}                      & \textbf{ClaimGen-CN} & \textbf{ClaimGen-CN-test} \\ \midrule
                        \midrule
\# Cause of Action                    & 100                &   100             \\
\# Sample                          &     207,748          &  1,000        \\
Avg. \# Fact Tokens             &   353.4    &  379.2    \\
Avg. \# Claim Tokens             &   134.3    &  120.1   \\\bottomrule
\end{tabular}}
\caption{Statistics of datasets.}
\label{tab:stat}
\end{table}

\begin{table*}[ht]
\small
\centering
\resizebox{\linewidth}{!}{
\begin{tabular}{lcccccc}
\toprule[2pt]
\textbf{Dataset}  & \textbf{\# Samples} & \textbf{\# Causes}  & \textbf{Avg. \# Claim Tokens}  & \textbf{Avg. \# Fact Tokens}  & \textbf{Availability} & \textbf{Plaintiff-centered}\\\midrule
                        \midrule
AutoJudge \citep{long2019automatic} & 100,000  & 1                  & 23.9                    & 100.1                 & \ding{52}  & \ding{56}       \\
AC-NLG \citep{wu2020biased}    & 66,904    & 1                  & 77.9                     & 158                    & \ding{52}   & \ding{56}      \\
MSJudge \citep{ma2021legal}   & 70,482    & 1                  & –  & 143                    & \ding{52}     & \ding{56}     \\
LK \citep{gan2021judgment}        & 61,611    & 1                  & –    & – & \ding{52}  & \ding{56}       \\
CCJudge \citep{zhao2021legal}   & 123,048   & 238                & 109.8                    & 198.7                  & \ding{56}  & \ding{56}       \\
CPEE \citep{zhao2022cpee}      & 158,625  & 10                 & 99                       & 156                    & \ding{52}    & \ding{56}      \\
ClaimGen-CN (Ours)  & 207,748   & 100                & 134.3                    & 353.4                  & \ding{52}     & \ding{52}  \\\bottomrule 
\end{tabular}}
\caption{Comparison of legal datasets. – means the value is not accessible.}
\label{tab:comparison}
\end{table*}

\begin{table}[ht]
\centering
\small
\resizebox{\linewidth}{!}{
    \begin{tabular}{lc}
    \toprule[2pt]
        \textbf{Causes of Action} & \textbf{Number} \\  \midrule \midrule
        Private Lending Disputes & 37,781 \\ 
        Motor Vehicle Traffic Accident Liability Disputes & 24,025 \\ 
        Sales Contract Disputes & 17,487 \\ 
        Financial Loan Contract Disputes & 13,231 \\ 
        Contract Disputes (General) & 8,906 \\ 
        ... & ... \\ 
        Catering Service Contract Disputes & 589 \\ 
        Trademark Infringement Disputes & 66 \\ 
        Child Support Disputes in Cohabitation Relationships & 70 \\ 
        Bank Card Disputes & 80 \\ 
        Child Support Disputes & 39 \\ \bottomrule
    \end{tabular}}
    \caption{Cause of action distribution.}
    \label{tab:cause_dis}    
\end{table}

\begin{table}[ht]
\centering
\small
\resizebox{\linewidth}{!}{
    \begin{tabular}{lc}
    \toprule[2pt]
        \textbf{Court} & \textbf{Number} \\ \midrule \midrule
        Shanghai Pudong New District People's Court & 3,452 \\ 
        Shanghai Minhang District People's Court & 2,329 \\ 
        Shanghai Baoshan District People's Court & 1,620 \\ 
        Shanghai Jing'an District People's Court & 1,470 \\ 
        Shanghai Qingpu District People's Court & 1,390 \\ 
        ... & ... \\
        Ningbo Maritime Court & 1 \\
        Yueyang Intermediate People's Court, Hunan Province & 1 \\
        Nantong Intermediate People's Court, Jiangsu Province & 1 \\ 
        Lixian People's Court & 1 \\
        Ankang Intermediate People's Court, Shaanxi Province & 1 \\ \bottomrule
    \end{tabular}}
    \caption{Court distribution.}
    \label{tab:court_dis}       
\end{table}

\begin{table}[ht]
\centering
\small
    \begin{tabular}{lc}
        \toprule[2pt]
        \textbf{Year} & \textbf{Number} \\ \midrule \midrule
        2010 & 7 \\
        2012 & 1 \\
        2013 & 3 \\
        2014 & 7 \\
        2015 & 21 \\
        2016 & 12,720 \\
        2017 & 47,749 \\
        2018 & 47,885 \\
        2019 & 38,966 \\
        2020 & 26,160 \\
        2021 & 28,656 \\
        2022 & 5,573 \\
        \bottomrule
    \end{tabular}
    \caption{Case start year distribution.}
    \label{tab:year_dis}          
\end{table}

\begin{table}[ht]
\centering
\small
\resizebox{\linewidth}{!}{
    \begin{tabular}{lccc}
    \toprule[2pt]
        \textbf{Statistic} & \textbf{Fact Length} & \textbf{Request Length} & \textbf{Verdict Length} \\ \midrule \midrule
        Mean & 353.4 & 134.3 & 605.8 \\ 
        Std Dev & 226.1 & 84.3 & 463.8 \\ 
        Min & 7.0 & 14.0 & 103.0 \\ 
        Max & 7,350.0 & 5,724.0 & 36,593.0 \\ \bottomrule
    \end{tabular}}
    \caption{Text length distribution.}
    \label{tab:text_dis}        
\end{table}

\subsection{Dataset Characteristics}

The comparison of ClaimGen-CN with other open-access legal datasets is presented in Table \ref{tab:comparison}. We discuss the characteristics of ClaimGen-CN based on the following aspects.

\textbf{Diverse.} Prior research mainly centered on private lending disputes, the most common cause of action. Among the existing open-source datasets, ClaimGen-CN stands out as the only one with more than 10 distinct civil case categories. ClaimGen-CN encompasses and openly shares a diverse range of case categories that were absent in prior work, including private lending disputes, divorce disputes, sales contract disputes, labor disputes, residential lease contract disputes, maintenance disputes, education and training contract disputes, and more.

\textbf{Large-scale.} As Table \ref{tab:comparison} illustrates, ClaimGen-CN is currently the largest civil litigation dataset, comprising 207k records and covering a wide variety of case types. This diversity in data provides researchers with numerous options for drafting legal claims. The number of tokens in the claims and facts in this dataset is much higher than in the previous datasets, which also introduces greater complexity.

\textbf{Plaintiff-centered.} ClaimGen-CN is a plaintiff-centered dataset that emphasizes the demands of plaintiffs. This dataset primarily focuses on the connection between the plaintiff's facts and their claims. Additionally, it shifts the focus from supporting judges to helping the non-expert public whose rights have been violated. 

\textbf{Comprehensive.} Beyond the general comparison, we further provide detailed distributions of ClaimGen-CN to demonstrate its comprehensiveness. As shown in Table~\ref{tab:cause_dis}, private lending disputes constitute the largest share, followed by motor vehicle traffic accident liability disputes and sales contract disputes, while a number of relatively rare categories such as trademark infringement and child support disputes also appear. Table~\ref{tab:court_dis} illustrates the court distribution, where cases are concentrated in Shanghai courts such as Pudong New District and Minhang District People’s Courts, though a small number of judgments come from courts in other provinces. Regarding temporal coverage, Table~\ref{tab:year_dis} shows that the dataset spans from 2010 to 2022, with a sharp increase starting in 2016 and peaking around 2017–2018 before declining in recent years. Finally, Table~\ref{tab:text_dis} reports the distribution of fact, request, and verdict text lengths, reflecting significant heterogeneity across different sections of judgments.

\section{Experiments}

\subsection{Baseline Models}

We implement the following baseline models for comparison\footnote{To our best knowledge, we are the first to propose the claim generation task, and there are no dedicated previous methods that could be directly applied for this task.}:
\textbf{GPT-4o} \citep{hurst2024gpt} is an optimized version of GPT-4 \citep{achiam2023gpt}, featuring superior natural language processing capabilities and multimodal interaction functions. It can handle various data types including text, images, and audio, and is suitable for a wide range of complex tasks and application scenarios.
\textbf{LLaMA3.1} \citep{grattafiori2024llama} possesses powerful multilingual dialogue and code generation capabilities after pre-training and instruction fine-tuning. 
\textbf{Claude3.5} \citep{anthropic2024claude35addendum} has enhanced performance, accuracy, and excels at high-level understanding and reasoning, providing reliable and consistent performance.
\textbf{Qwen2.5} \citep{qwen2.5} has strong instruction following ability, improved coding and mathematical ability, can generate long text, understand structured data, and generate structured output.
\textbf{Deepseek-R1} \citep{guo2025deepseek} is efficient and flexible, and performs well on reasoning, coding, and mathematics tasks with lower training cost. At the same time, it can be applied to various application scenarios that require multi-modal processing such as image description generation.
\textbf{Farui} \citep{alibaba2024farui} is a large-model-based AI legal advisor trained on professional legal data, capable of legal knowledge understanding, reasoning, and generation.

We apply 0-shot settings to GPT-4o \citep{hurst2024gpt}, LLaMA3.1 \citep{grattafiori2024llama}, Claude3.5 \citep{anthropic2024claude35addendum}, Qwen2.5 \citep{qwen2.5}, DeepSeek-R1 \citep{guo2025deepseek} and Farui \citep{alibaba2024farui}. Details of LLMs are provided in Appendix \ref{app:llm}.

\subsection{Experiment Settings}

Here we describe the implementation of the claim generation method used in our experiments. Note that all LLMs are replaceable in this method. For all models (GPT-4o, Claude 3.5, DeepSeek-R1, Farui, LLaMA3.1, and Qwen2.5), we used the official APIs or open-source checkpoints available via OpenAI, Anthropic, Deepseek, Aliyun, or Hugging Face. See Appendix~\ref{app:urls} for full API URLs.

We adopt a zero-shot setting for all models evaluated in this study. Each model is prompted to generate the plaintiff’s legal claims based solely on the factual description of a given case. We provide the full factual context as input to the model, without any additional fine-tuning, demonstrations, or retrieval-based augmentation. The prompt used across all models is as follows: \textit{``Please generate the plaintiff's claims based on the following facts.''}

\subsection{Metrics}

When constructing our dataset, we have considered including common NLG metrics such as ROUGE \citep{lin-2004-rouge}, BLEU \citep{papineni2002bleu}, and BERT SCORE \citep{zhang2019bertscore} to help users evaluate text quality. However, these metrics do not provide a good measure of the overall claim quality \citep{DBLP:conf/emnlp/LiuLSNCP16,DBLP:conf/emnlp/NovikovaDCR17,chaganty2018price,DBLP:conf/acl/SellamDP20,DBLP:conf/conll/DeutschR21,liu2023llms}. To better assess claim quality, we introduce two metrics: factuality and clarity for evaluating the claims. We employ LLM, specifically GPT-4o \citep{hurst2024gpt}, for scoring. GPT-4o evaluates claims using defined prompts for these metrics, as detailed in Appendix \ref{sec:GPT4ScoringPrompts}. The robustness and reliability of our proposed metrics are detailed in Appendix \ref{sec:MetaEvaluation}.

\textbf{ROUGE} \citep{lin-2004-rouge} is a set of metrics used in the NLP task. We keep the results of ROUGE-1, ROUGE-2, and ROUGE-L. ROUGE-1 and ROUGE-2 refer to the overlap of unigram and bigram between the generated and reference documents, respectively. ROUGE-L is a Longest Common Subsequence (LCS) based statistics.

\textbf{BLEU} \citep{papineni2002bleu} is used for automatic text-generation evaluation and correlates well with human judgment. We evaluate using BLEU-1, BLEU-2, and average scores from BLEU-1 through BLEU-4.

\textbf{BERT SCORE} \citep{zhang2019bertscore} measures similarity using contextual embeddings. We calculate the precision (p), recall (r), and f1-score to evaluate the information matching degree.

\textbf{Factuality} refers to the facts stated in the claims being truthful, accurate, and based on objectively existing circumstances. 

\textbf{Clarity} means that claims should be specific, providing details such as the amount of compensation for losses and specifying the manner and scope of issuing an apology.

\textbf{Total} represents the average of factuality and clarity scores.

\begin{table*}[ht]
\small
    \centering
\begin{tabular}{lccc|ccc|ccc}
\toprule[2pt]
\multirow{2}{*}{\textbf{Method}} & \multicolumn{3}{c|}{\textbf{BLEU}} & \multicolumn{3}{c|}{\textbf{ROUGE}} & \multicolumn{3}{c}{\textbf{BERT   SCORE}} \\
                        & B-1     & B-2    & B-N   & R-1     & R-2    & R-L    & p         & r         & f1   \\ \midrule
                        \midrule
GPT-4o \citep{hurst2024gpt}                   & 19.07   & 7.25   & 7.85  & \underline{48.98}   & 22.40  & \textbf{43.78}  & 76.06     & 65.86     & 70.46   \\
LLaMA3.1 \citep{grattafiori2024llama}    & 16.95   & 5.64   & 6.82  & 43.14   & 19.12  & 36.05  & 69.38     & \textbf{70.58}     & 69.64   \\
Claude3.5 \citep{anthropic2024claude35addendum}       & \textbf{22.07}   & \textbf{9.59}   & \textbf{9.92}  & \textbf{49.26}   & \textbf{24.21}  & \underline{42.95}  & 76.56     & \underline{67.70}     & \textbf{71.66}    \\
Qwen2.5 \citep{qwen2.5}     & 18.59   & 7.81   & 8.22  & 46.26   & 22.10  & 41.00  & \textbf{77.76}     & 64.46     & 70.34    \\
DeepSeek-R1  \citep{guo2025deepseek}               & 14.16   & 5.61   & 5.99  & 42.32   & 18.29  & 37.38  & \underline{77.59}     & 60.56     & 67.91    \\
Farui \citep{alibaba2024farui}                    & \underline{20.81}   & \underline{8.54}   & \underline{9.08}  & 48.60   & \underline{23.10}  & 42.55  & 76.28     & 66.45     & \underline{70.85}    \\
\bottomrule
\end{tabular}
\caption{Automatic evaluation results on the \textsc{ClaimGen-CN-test} set using ROUGE, BLEU, and BERT SCORE. The best is \textbf{bolded} and the second best is \underline{underlined}.}
\label{tab:auto_eval}
\end{table*}

\begin{table*}[ht]
\small
    \centering
\begin{tabular}{lccc}
\toprule[2pt]
\multirow{2}{*}{\textbf{Method}} & \multicolumn{3}{c}{\textbf{GPT-4o   SCORE}} \\
\multicolumn{1}{c}{}                        &  Factuality   & Clarity  & Total  \\ \midrule
                        \midrule
GPT-4o \citep{hurst2024gpt}                  &  48.31        & 56.23    & 52.27  \\
LLaMA3.1 \citep{grattafiori2024llama}    & 51.32        & 55.18    & 53.25  \\
Claude3.5 \citep{anthropic2024claude35addendum}        & 54.18        & \underline{64.16}    & \underline{59.17}  \\
Qwen2.5 \citep{qwen2.5}       & \underline{54.34}        & 59.61    & 56.97  \\
DeepSeek-R1 \citep{guo2025deepseek}                & \textbf{62.14}        & \textbf{69.43}    & \textbf{65.79}  \\
Farui \citep{alibaba2024farui}                     & 42.85        & 46.28    & 44.56  \\
\bottomrule
\end{tabular}
\caption{Evaluation results on the \textsc{ClaimGen-CN-test} set using GPT-4o. Total represents the average of factuality and clarity scores. The best is \textbf{bolded} and the second best is \underline{underlined}.}
\label{tab:gpt4_eval}
\end{table*}

\subsection{Experimental Results}

We report the performance of several state-of-the-art models on the \textsc{ClaimGen-CN-test} set using both automatic and human-aligned evaluations. The automatic metrics include BLEU, ROUGE, and BERT SCORE (Table~\ref{tab:auto_eval}), while GPT-4o-based human-aligned scores focus on factuality and clarity (Table~\ref{tab:gpt4_eval}).

\paragraph{Automatic Evaluation.}
As shown in Table~\ref{tab:auto_eval}, Claude3.5 achieves the highest performance across most automatic metrics, including BLEU, ROUGE, and BERT SCORE f1. Farui ranks second in BLEU and BERT SCORE f1, while GPT-4o leads in ROUGE-L but is outperformed by Claude3.5 and Qwen2.5 on other metrics. These results suggest that Claude3.5 generates outputs with higher n-gram overlap and semantic similarity to the references, as captured by both lexical and contextual metrics.

\paragraph{GPT-4o Evaluation.}
To better understand the alignment between generated claims and human judgments, we use GPT-4o to evaluate outputs based on \textbf{Factuality} and \textbf{Clarity} (Table~\ref{tab:gpt4_eval}). DeepSeek-R1 significantly outperforms other models with a total score of 65.79, showing the best factual accuracy and clarity. Claude3.5 again demonstrates strong performance with the second-best total score, while Qwen2.5 and LLaMA3.1 achieve moderate scores. While GPT-4o performs well on automatic metrics, its scores in GPT-4o-based evaluations are lower than those of several other models.

Overall, Claude3.5 stands out in terms of automatic metrics, whereas DeepSeek-R1 excels in GPT-4o evaluation. This discrepancy underscores the importance of incorporating human-centric evaluation to complement automatic metrics, especially for complex legal text generation tasks.

\subsection{Human Evaluation}
\label{sec:human-eval}
To assess the alignment between GPT-4o evaluations and human judgments, we calculate the mean absolute error (MAE) and consistency score between GPT-4o and human annotations over 100 randomly sampled cases generated by DeepSeek-R1. We adopt \textbf{Factuality} and \textbf{Clarity} as evaluation dimensions. Three annotators are asked to assign scores on a 1-to-5 scale, where 1 denotes the worst and 5 the best. 

\begin{table}[htbp]
\small
\centering
\begin{tabular}{lcc}
\toprule[2pt]
\textbf{Dimension} & \textbf{MAE}~$\downarrow$ & \textbf{Consistency}~$\uparrow$ \\
\midrule
                        \midrule
Factuality & 0.19 & 81.05 \\
Clarity    & 0.20 & 73.68 \\
\bottomrule
\end{tabular}
\caption{MAE and consistency between GPT-4o scores and human annotations on 100 DeepSeek-generated samples.}
\label{tab:mae_consistency}
\end{table}

As shown in Table~\ref{tab:mae_consistency}, GPT-4o achieves a MAE of 0.19 and a consistency of 81.05 on the factuality dimension, and a MAE of 0.20 with 73.68 consistency on clarity. These results suggest that GPT-4o evaluations are closely aligned with human judgments, as evidenced by the low MAE and high consistency scores across both dimensions. In particular, the model demonstrates stronger agreement with human ratings on factuality, achieving a consistency of 81.05, compared to 73.68 on clarity. This indicates that while GPT-4o can serve as a trustworthy proxy for human annotation overall, its performance is more stable and reliable in assessing factual correctness than in evaluating clarity, where subjectivity may play a greater role.

A detailed analysis of human evaluation is provided in Appendix~\ref{sec:annotation-guidelines}, Appendix~\ref{sec:annotation-protocol} and Appendix~\ref{sec:HumanEvaluation}.

\subsection{Case Study}
\begin{figure*}[ht]
    \centering
    \includegraphics[width=1\textwidth]{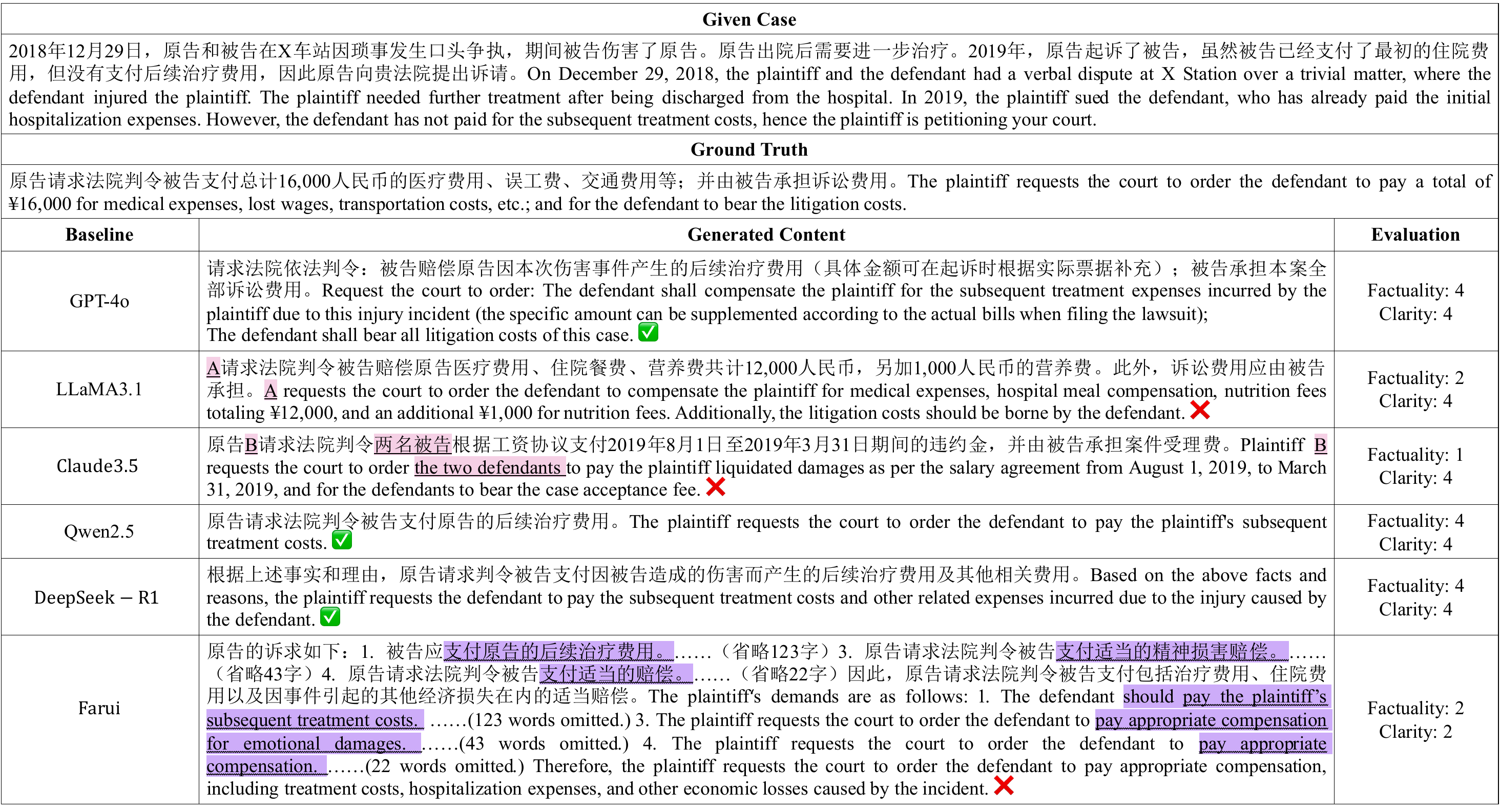}
    \caption{The claim generation of a given case. The \textcolor{mypurple}{\uwave{purple}} parts indicate clarity mistakes in prediction, while the \textcolor{darkpink}{\underline{pink}} parts are factual mistakes in prediction. Green checkmarks indicate acceptable outputs, while red crosses mark outputs with factual or clarity errors.}
    \vspace{-10pt}
    \label{fig:showcase}
\end{figure*}

Figure~\ref{fig:showcase} illustrates a representative case from the ClaimGen-CN-test set, showcasing the comparative outputs of different large language models (LLMs) when generating legal claims. The factual background involves a plaintiff injured in a dispute, seeking compensation for subsequent treatment costs not covered by the defendant.

The ground truth specifies that the plaintiff requests the defendant to pay a total of ¥16,000 covering medical expenses, lost wages, transportation costs, and litigation expenses. Among the models evaluated, GPT-4o, Qwen2.5, and DeepSeek-R1 generate legally accurate and contextually faithful claims, aligning well with the facts and compensation details. In particular, GPT-4o excels in both factuality and clarity, offering a detailed yet concise articulation of the plaintiff’s request.

In contrast, models like Claude3.5 and LLaMA3.1 either misidentify the defendants or fabricate salary agreements, resulting in factual inconsistencies. Farui's output, although structurally rich, suffers from vague and overly general phrasing, leading to reduced clarity and a failure to directly match the required compensation scope.

This case highlights the importance of grounding legal claim generation not only in accurate fact extraction but also in the precise mapping of legal entitlements, where minor deviations can significantly impact the validity of generated content.

\section{Discussion}
\subsection{Error Analysis}
\label{sec:error-analysis}
Current models exhibit multidimensional deficiencies in legal document generation, primarily manifested through four core aspects: inadequate legal knowledge comprehension, disconnects in legal-mathematical logic, polarized deviations in claim generation, and systemic instability in output precision.

The following case examples are provided in Appendix~\ref{sec:cases} to supplement the error analysis discussed in this section. We include full input facts, ground truth and model outputs for all referenced instances: \textit{CaseID~1, 2, 3, 11, 93, 98, 110, and 116}, in the order of appearance.

\textbf{Lack of Legal Knowledge.} We identify two common types of legal knowledge gaps that cause models to generate incorrect or incomplete claims. First, models often fail to reconstruct legally relevant facts from the case description. In \textit{CaseID~1}, for instance, the model mistakenly assumed that interest began accruing a month too early. This shows the lack of ability to track and reason over event timelines based on contractual terms and payment records—an essential skill for accurate legal fact recognition. Second, models lack domain-specific legal knowledge needed for handling different types of cases. For example, in loan disputes, the validity of a claim often depends on understanding the sequence of events—loan issuance, due date, default, and interest—along with the corresponding legal rules. Without knowledge of such patterns, the model may generate structurally invalid or legally irrelevant claims.

\textbf{Legal-Mathematical Disconnects.}
LLMs often struggle with multi-step quantitative legal reasoning. In \textit{CaseID~2}, for example, most models failed to calculate the correct inheritance share in a co-ownership scenario. The legally precise logic—$50\% \times \frac{1}{4} = \frac{1}{8}$-was reduced to vague statements like ``proportional division,'' with no explanation of how the entitlement was derived. This error points to a deeper gap between legal texts and computational understanding. Specifically, models lack the ability to interpret and apply statutory provisions such as Articles 1122 and 1130 of the Civil Code of the People’s Republic of China, which define the structure and sequence of inheritance allocation. As a result, they fail to generate legally valid and numerically grounded claims.

\textbf{Polarized Claim Generation.}
Models exhibit two opposite types of errors when generating claims: some produce redundant requests that are not supported by the facts, while others omit essential claims required by the legal context. In \textit{CaseID~3} and \textit{CaseID~11}, GPT-4o and Qwen2.5 inserted mental damage compensation and interest claims that the plaintiff never requested, likely reflecting patterns overrepresented in debt-related training examples. On the other hand, Qwen2.5 and LLaMA3.1 omitted essential claims such as the confirmation of contract validity in \textit{CaseID~11}, which is a legal prerequisite for pursuing the associated property transfer. These errors reflect two distinct but co-existing problems. First, some models tend to hallucinate legally implausible claims by defaulting to common case patterns rather than the actual fact statements. Second, others fail to identify necessary legal elements that are implied by the factual background or logically required by the cause of action. Consequently, claims that are either excessive or incomplete are produced, which reduces the factual and legal accuracy of the output.

\textbf{Systemic Instability.}
LLaMA3.1 shows repeated output problems in two ways: (1) In \textit{CaseID~93}, it copied and pasted over 180 consumer protection law articles without selecting the relevant ones, making the response bloated and meaningless;
(2) In \textit{CaseID~98, 110, and 116}, LLaMA3.1 repeated similar legal claims more than five times, showing a lack of control over output length and content. In practical terms, this hurts the clarity and coherence of generated claims, making it difficult for users to identify distinct legal requests. More importantly, these repeated and redundant outputs obscure the logical structure of the plaintiff’s position, potentially causing critical claims to be overlooked amid excessive or noisy output. This undermines the functional purpose of legal claim generation as a concise, actionable summary of the plaintiff’s demands.

\subsection{Future Work in Legal Claim Generation}
Our error analysis reveals deeper challenges in legal claim generation, pointing to several scenario-driven directions for future research. First, large–small model collaboration can improve factual grounding, such as using lightweight modules to identify key events or legal rules before invoking a larger model for structured claim generation. Second, long-chain reasoning techniques may help track complex legal timelines—e.g., loan issuance, default, and interest accrual—enhancing logical completeness. Third, reinforcement learning with legal-specific feedback can be used to optimize the claim generation process by designing task-specific reward functions. For instance, models can be penalized for producing claims that lack factual support, and rewarded for generating claims that follow legally valid reasoning paths. Such techniques should be developed in close alignment with real-world legal tasks to ensure reliability and practical value.

Although our dataset is constructed based on the Chinese legal system, the overall data construction workflow and the evaluation design are not specific to China. With appropriate adjustments, especially considering the role of precedents in common law countries, our approach can be applied to other legal systems as well.

\section{Conclusion}
In this paper, we make a pioneering step in Legal AI by focusing on generating claims for civil litigation. Our dataset, ClaimGen-CN, covers a wide range of legal cases, offering a solid foundation for future Legal AI research and application. To evaluate model performance, we propose two dimensions, namely factuality and clarity, which are tailored to the specific requirements of legal text generation. Our error analysis reveals that current LLMs struggle with generating factually accurate
and concise claims, suggesting key directions for future improvements.

\section*{Limitations}
In this section, we discuss the limitations of our work:

$\bullet$ We only interact with the LLMs one round per time. The LLMs are capable of multi-round interaction, which may help the LLMs to better understand the claim generation task.

$\bullet$ The reliance on LLMs for claim generation raises questions about the transparency and interpretability of the generated claims. The decision-making process of these models is often opaque, which could lead to challenges in understanding and justifying the basis on which claims are generated.

$\bullet$ The exploration of legal claim generation tasks beyond the Chinese legal context remains limited. 

\section*{Ethics Statement}
The corpus we use is released by the Chinese government and has been anonymized wherever necessary\footnote{Provisions of the Supreme People's Court on Publication of Judgment Documents by the People's Courts on the Internet 2016. \url{http://gongbao.court.gov.cn/Details/415f49dd8baaa04b479d57af9616ef.html}}. Our dataset thus does not involve any personally private information. Besides, the corpus is in public domain and licensed for use within the legal scope\footnote{See Article 3.3 of User Protocol of China Judgments Online. \url{https://g.alicdn.com/onlineCourt/static/0.6.97/akan-wenshu-protocol.html}}. Specifically, sensitive personal identifiers (such as names, identification numbers, and contact details) were removed or masked by the data providers before public release. Our research strictly used these desensitized datasets and did not involve any additional processing that could lead to re-identification. Beyond relying on officially anonymized and desensitized documents, we emphasize that our research activities did not introduce or generate any new sensitive information.

With the increasing use of Legal AI for justice-related tasks, there is growing concern about its ethical implications, including the risk of biases and errors that could have serious repercussions.
To address these concerns, it's important to clarify that our work is an exploration of a new data source and the development of an algorithm, which is not meant to be immediately and directly used in practical settings. Our goal is to assist non-professionals by providing recommendations, not to make final decisions.

\textbf{Measures for releasing the dataset:} The dataset was solely used for academic purposes such as model evaluation and error analysis. No attempts were made to re-identify individuals, and no system or output of this work has been, or will be, deployed in real-world legal practice. We emphasize that AI systems in legal contexts must be subject to strict human oversight, and our work is intended to support research and understanding rather than to replace human judgment. Any potential application of similar technologies should therefore incorporate comprehensive oversight mechanisms to prevent misuse.

\textbf{Measures for future use:} The proposed framework will include warning statements, such as adding the phrase “This answer may be incorrect and is for reference only” in the output. We will also insert declarations in the prompt instructing the model not to provide answers when confidence is low or requirements are not met, and we will consider additional validation after generation. The proposed framework will also advise users to seek professional legal advice, for instance, adding “Before making a final decision, you should consult a qualified lawyer” in the output.

\section*{Acknowledgements}

We would like to thank the anonymous reviewers and the Area Chair for their constructive feedback and insightful suggestions. We are also sincerely thankful to Chen Yanwen for suggesting the name of the dataset and for providing valuable comments and revisions.

\normalem
\bibliography{anthology,custom}

\begin{thebibliography}{53}
\providecommand{\natexlab}[1]{#1}

\bibitem[{Achiam et~al.(2023)Achiam, Adler, Agarwal, Ahmad, Akkaya, Aleman, Almeida, Altenschmidt, Altman, Anadkat et~al.}]{achiam2023gpt}
Josh Achiam, Steven Adler, Sandhini Agarwal, Lama Ahmad, Ilge Akkaya, Florencia~Leoni Aleman, Diogo Almeida, Janko Altenschmidt, Sam Altman, Shyamal Anadkat, and 1 others. 2023.
\newblock \href {https://doi.org/10.48550/arXiv.2303.08774} {{GPT-4} technical report}.
\newblock \emph{arXiv preprint arXiv:2303.08774}.

\bibitem[{Aletras et~al.(2016)Aletras, Tsarapatsanis, Preo{\c{t}}iuc-Pietro, and Lampos}]{aletras2016predicting}
Nikolaos Aletras, Dimitrios Tsarapatsanis, Daniel Preo{\c{t}}iuc-Pietro, and Vasileios Lampos. 2016.
\newblock \href {https://doi.org/10.7717/PEERJ-CS.93} {Predicting judicial decisions of the european court of human rights: A natural language processing perspective}.
\newblock \emph{PeerJ computer science}, 2:e93.

\bibitem[{{Alibaba Cloud}(2024)}]{alibaba2024farui}
{Alibaba Cloud}. 2024.
\newblock \href {https://help.aliyun.com/zh/model-studio/tongyi-farui-api} {Tongyi farui: Legal large language model}.
\newblock Accessed May 18, 2025.

\bibitem[{Anthropic(2024)}]{anthropic2024claude35addendum}
Anthropic. 2024.
\newblock Claude 3.5 sonnet model card addendum.
\newblock \url{https://www-cdn.anthropic.com/fed9cc193a14b84131812372d8d5857f8f304c52/Model_Card_Claude_3_Addendum.pdf}.
\newblock Accessed May 18, 2025.

\bibitem[{Bhattacharya et~al.(2020)Bhattacharya, Ghosh, Pal, and Ghosh}]{bhattacharya2020hier}
Paheli Bhattacharya, Kripabandhu Ghosh, Arindam Pal, and Saptarshi Ghosh. 2020.
\newblock \href {https://doi.org/10.1145/3397271.3401191} {Hier-spcnet: a legal statute hierarchy-based heterogeneous network for computing legal case document similarity}.
\newblock In \emph{Proceedings of the 43rd international ACM SIGIR conference on research and development in information retrieval}, pages 1657--1660.

\bibitem[{Bommarito et~al.(2018)Bommarito, Katz, and Detterman}]{bommarito2018lexnlp}
MJ~Bommarito, Daniel~Martin Katz, and E~Detterman. 2018.
\newblock \href {https://doi.org/10.48550/arXiv.1806.03688} {Lexnlp: Natural language processing and information extraction for legal and regulatory texts}.
\newblock \emph{Research Handbook on Big Data Law}.

\bibitem[{Cardellino et~al.(2017)Cardellino, Teruel, Alemany, and Villata}]{cardellino2017legal}
Cristian Cardellino, Milagro Teruel, Laura~Alonso Alemany, and Serena Villata. 2017.
\newblock \href {https://doi.org/10.18653/v1/e17-2041} {Legal {NERC} with ontologies, wikipedia and curriculum learning}.
\newblock In \emph{Proceedings of the 15th Conference of the European Chapter of the Association for Computational Linguistics, {EACL} 2017, Valencia, Spain, April 3-7, 2017, Volume 2: Short Papers}, pages 254--259. Association for Computational Linguistics.

\bibitem[{Chaganty et~al.(2018)Chaganty, Mussmann, and Liang}]{chaganty2018price}
Arun Chaganty, Stephen Mussmann, and Percy Liang. 2018.
\newblock The price of debiasing automatic metrics in natural language evalaution.
\newblock In \emph{Proceedings of the 56th Annual Meeting of the Association for Computational Linguistics (Volume 1: Long Papers)}, pages 643--653.

\bibitem[{Chalkidis et~al.(2020)Chalkidis, Fergadiotis, Malakasiotis, Aletras, and Androutsopoulos}]{chalkidis2020legal}
Ilias Chalkidis, Manos Fergadiotis, Prodromos Malakasiotis, Nikolaos Aletras, and Ion Androutsopoulos. 2020.
\newblock \href {https://arxiv.org/abs/2010.02559} {{LEGAL-BERT:} the muppets straight out of law school}.
\newblock \emph{CoRR}, abs/2010.02559.

\bibitem[{Chalkidis et~al.(2022)Chalkidis, Jana, Hartung, Bommarito, Androutsopoulos, Katz, and Aletras}]{chalkidis2021lexglue}
Ilias Chalkidis, Abhik Jana, Dirk Hartung, Michael Bommarito, Ion Androutsopoulos, Daniel Katz, and Nikolaos Aletras. 2022.
\newblock \href {https://doi.org/10.18653/v1/2022.acl-long.297} {{L}ex{GLUE}: A benchmark dataset for legal language understanding in {E}nglish}.
\newblock In \emph{Proceedings of the 60th Annual Meeting of the Association for Computational Linguistics (Volume 1: Long Papers)}, pages 4310--4330, Dublin, Ireland. Association for Computational Linguistics.

\bibitem[{Chien et~al.(2024)Chien, Chang, and Sun}]{chien2024legal}
Kuo-Chun Chien, Chia-Hui Chang, and Ren-Der Sun. 2024.
\newblock \href {https://doi.org/10.1016/j.clsr.2023.105902} {Legal knowledge management for prosecutors based on judgment prediction and error analysis from indictments}.
\newblock \emph{Computer Law \& Security Review}, 52:105902.

\bibitem[{Deutsch and Roth(2021)}]{DBLP:conf/conll/DeutschR21}
Daniel Deutsch and Dan Roth. 2021.
\newblock \href {https://doi.org/10.18653/V1/2021.CONLL-1.24} {Understanding the extent to which content quality metrics measure the information quality of summaries}.
\newblock In \emph{Proceedings of the 25th Conference on Computational Natural Language Learning, CoNLL 2021, Online, November 10-11, 2021}, pages 300--309. Association for Computational Linguistics.

\bibitem[{Fabbri et~al.(2021)Fabbri, Kryscinski, McCann, Xiong, Socher, and Radev}]{DBLP:journals/tacl/FabbriKMXSR21}
Alexander~R. Fabbri, Wojciech Kryscinski, Bryan McCann, Caiming Xiong, Richard Socher, and Dragomir~R. Radev. 2021.
\newblock \href {https://doi.org/10.1162/TACL\_A\_00373} {Summeval: Re-evaluating summarization evaluation}.
\newblock \emph{Trans. Assoc. Comput. Linguistics}, 9:391--409.

\bibitem[{Feng et~al.(2022)Feng, Li, and Ng}]{feng2022legal}
Yi~Feng, Chuanyi Li, and Vincent Ng. 2022.
\newblock \href {https://doi.org/10.18653/V1/2022.ACL-LONG.48} {Legal judgment prediction via event extraction with constraints}.
\newblock In \emph{Proceedings of the 60th Annual Meeting of the Association for Computational Linguistics (Volume 1: Long Papers)}, pages 648--664.

\bibitem[{Gan et~al.(2021)Gan, Kuang, Yang, and Wu}]{gan2021judgment}
Leilei Gan, Kun Kuang, Yi~Yang, and Fei Wu. 2021.
\newblock \href {https://doi.org/10.1609/AAAI.V35I14.17522} {Judgment prediction via injecting legal knowledge into neural networks}.
\newblock In \emph{Thirty-Fifth {AAAI} Conference on Artificial Intelligence, {AAAI} 2021, Thirty-Third Conference on Innovative Applications of Artificial Intelligence, {IAAI} 2021, The Eleventh Symposium on Educational Advances in Artificial Intelligence, {EAAI} 2021, Virtual Event, February 2-9, 2021}, pages 12866--12874. {AAAI} Press.

\bibitem[{Grattafiori et~al.(2024)Grattafiori, Dubey, Jauhri, Pandey, Kadian, Al-Dahle, Letman, Mathur, Schelten, Vaughan et~al.}]{grattafiori2024llama}
Aaron Grattafiori, Abhimanyu Dubey, Abhinav Jauhri, Abhinav Pandey, Abhishek Kadian, Ahmad Al-Dahle, Aiesha Letman, Akhil Mathur, Alan Schelten, Alex Vaughan, and 1 others. 2024.
\newblock The llama 3 herd of models.
\newblock \emph{arXiv preprint arXiv:2407.21783}.

\bibitem[{Guo et~al.(2025)Guo, Yang, Zhang, Song, Zhang, Xu, Zhu, Ma, Wang, Bi et~al.}]{guo2025deepseek}
Daya Guo, Dejian Yang, Haowei Zhang, Junxiao Song, Ruoyu Zhang, Runxin Xu, Qihao Zhu, Shirong Ma, Peiyi Wang, Xiao Bi, and 1 others. 2025.
\newblock Deepseek-r1: Incentivizing reasoning capability in llms via reinforcement learning.
\newblock \emph{arXiv preprint arXiv:2501.12948}.

\bibitem[{Hachey and Grover(2006)}]{hachey2006extractive}
Ben Hachey and Claire Grover. 2006.
\newblock \href {https://doi.org/10.1007/s10506-007-9039-z} {Extractive summarisation of legal texts}.
\newblock \emph{Artif. Intell. Law}, 14(4):305--345.

\bibitem[{He(2022)}]{he2022}
Haibo He. 2022.
\newblock \emph{Administrative Litigation Law (3rd Edition)}.
\newblock Law Press.

\bibitem[{Hendrycks et~al.(2021)Hendrycks, Burns, Chen, and Ball}]{hendrycks2021cuad}
Dan Hendrycks, Collin Burns, Anya Chen, and Spencer Ball. 2021.
\newblock \href {https://datasets-benchmarks-proceedings.neurips.cc/paper/2021/hash/6ea9ab1baa0efb9e19094440c317e21b-Abstract-round1.html} {{CUAD:} an expert-annotated {NLP} dataset for legal contract review}.
\newblock In \emph{Proceedings of the Neural Information Processing Systems Track on Datasets and Benchmarks 1, NeurIPS Datasets and Benchmarks 2021, December 2021, virtual}.

\bibitem[{Hurst et~al.(2024)Hurst, Lerer, Goucher, Perelman, Ramesh, Clark, Ostrow, Welihinda, Hayes, Radford et~al.}]{hurst2024gpt}
Aaron Hurst, Adam Lerer, Adam~P Goucher, Adam Perelman, Aditya Ramesh, Aidan Clark, AJ~Ostrow, Akila Welihinda, Alan Hayes, Alec Radford, and 1 others. 2024.
\newblock Gpt-4o system card.
\newblock \emph{arXiv preprint arXiv:2410.21276}.

\bibitem[{Katz et~al.(2023)Katz, Hartung, Gerlach, Jana, and Bommarito~II}]{katz2023natural}
Daniel~Martin Katz, Dirk Hartung, Lauritz Gerlach, Abhik Jana, and Michael~J Bommarito~II. 2023.
\newblock \href {https://doi.org/10.48550/ARXIV.2302.12039} {Natural language processing in the legal domain}.
\newblock \emph{arXiv preprint arXiv:2302.12039}.

\bibitem[{Krauss(2009)}]{krauss2009theory}
Rebecca Krauss. 2009.
\newblock The theory of prosecutorial discretion in federal law: Origins and developments.
\newblock \emph{Seton Hall Cir. Rev.}, 6:1.

\bibitem[{Landis and Koch(1977)}]{landis1977measurement}
J~Richard Landis and Gary~G Koch. 1977.
\newblock The measurement of observer agreement for categorical data.
\newblock \emph{biometrics}, pages 159--174.

\bibitem[{Le et~al.(2024)Le, Xiao, Xiao, and Li}]{le2024topology}
Yuquan Le, Sheng Xiao, Zheng Xiao, and Kenli Li. 2024.
\newblock \href {https://doi.org/10.1016/J.ESWA.2023.122103} {Topology-aware multi-task learning framework for civil case judgment prediction}.
\newblock \emph{Expert Systems with Applications}, 238:122103.

\bibitem[{Li et~al.(2025)Li, Wu, Cai, Jatowt, Zhou, Lu, Sun, Wu, and Kuang}]{li-etal-2025-legal}
Ang Li, Yiquan Wu, Ming Cai, Adam Jatowt, Xiang Zhou, Weiming Lu, Changlong Sun, Fei Wu, and Kun Kuang. 2025.
\newblock \href {https://aclanthology.org/2025.naacl-long.355/} {Legal judgment prediction based on knowledge-enhanced multi-task and multi-label text classification}.
\newblock In \emph{Proceedings of the 2025 Conference of the Nations of the Americas Chapter of the Association for Computational Linguistics: Human Language Technologies (Volume 1: Long Papers)}, pages 6957--6970, Albuquerque, New Mexico. Association for Computational Linguistics.

\bibitem[{Li(2023)}]{liWorldsSeeCuriosity2023}
Dr~Fei-Fei Li. 2023.
\newblock \emph{The {{Worlds I See}}: {{Curiosity}}, {{Exploration}}, and {{Discovery}} at the {{Dawn}} of {{AI}}}.
\newblock Flatiron Books: A Moment of Lift Book, New York.

\bibitem[{Lin(2004)}]{lin-2004-rouge}
Chin-Yew Lin. 2004.
\newblock \href {https://aclanthology.org/W04-1013} {{ROUGE}: A package for automatic evaluation of summaries}.
\newblock In \emph{Text Summarization Branches Out}, pages 74--81, Barcelona, Spain. Association for Computational Linguistics.

\bibitem[{Liu et~al.(2016)Liu, Lowe, Serban, Noseworthy, Charlin, and Pineau}]{DBLP:conf/emnlp/LiuLSNCP16}
Chia{-}Wei Liu, Ryan Lowe, Iulian Serban, Michael~D. Noseworthy, Laurent Charlin, and Joelle Pineau. 2016.
\newblock \href {https://doi.org/10.18653/V1/D16-1230} {How {NOT} to evaluate your dialogue system: An empirical study of unsupervised evaluation metrics for dialogue response generation}.
\newblock In \emph{Proceedings of the 2016 Conference on Empirical Methods in Natural Language Processing, {EMNLP} 2016, Austin, Texas, USA, November 1-4, 2016}, pages 2122--2132. The Association for Computational Linguistics.

\bibitem[{Liu et~al.(2023{\natexlab{a}})Liu, Wu, Zhang, Sun, Lu, Wu, and Kuang}]{liu2023ml}
Yifei Liu, Yiquan Wu, Yating Zhang, Changlong Sun, Weiming Lu, Fei Wu, and Kun Kuang. 2023{\natexlab{a}}.
\newblock \href {https://doi.org/10.1145/3539618.3591731} {{ML-LJP:} multi-law aware legal judgment prediction}.
\newblock In \emph{Proceedings of the 46th International ACM SIGIR Conference on Research and Development in Information Retrieval}, pages 1023--1034.

\bibitem[{Liu et~al.(2023{\natexlab{b}})Liu, Moosavi, and Lin}]{liu2023llms}
Yiqi Liu, Nafise~Sadat Moosavi, and Chenghua Lin. 2023{\natexlab{b}}.
\newblock \href {https://doi.org/10.48550/ARXIV.2311.09766} {Llms as narcissistic evaluators: When ego inflates evaluation scores}.
\newblock \emph{arXiv preprint arXiv:2311.09766}.

\bibitem[{Long et~al.(2019)Long, Tu, Liu, and Sun}]{long2019automatic}
Shangbang Long, Cunchao Tu, Zhiyuan Liu, and Maosong Sun. 2019.
\newblock \href {https://doi.org/10.1007/978-3-030-32381-3\_45} {Automatic judgment prediction via legal reading comprehension}.
\newblock In \emph{Chinese Computational Linguistics: 18th China National Conference, CCL 2019, Kunming, China, October 18--20, 2019, Proceedings 18}, pages 558--572. Springer.

\bibitem[{Ma et~al.(2021)Ma, Zhang, Wang, Liu, Ye, Sun, and Zhang}]{ma2021legal}
Luyao Ma, Yating Zhang, Tianyi Wang, Xiaozhong Liu, Wei Ye, Changlong Sun, and Shikun Zhang. 2021.
\newblock \href {https://doi.org/10.1145/3404835.3462945} {Legal judgment prediction with multi-stage case representation learning in the real court setting}.
\newblock In \emph{Proceedings of the 44th International ACM SIGIR Conference on Research and Development in Information Retrieval}, pages 993--1002.

\bibitem[{Malik et~al.(2021)Malik, Sanjay, Nigam, Ghosh, Guha, Bhattacharya, and Modi}]{malik-etal-2021-ildc}
Vijit Malik, Rishabh Sanjay, Shubham~Kumar Nigam, Kripabandhu Ghosh, Shouvik~Kumar Guha, Arnab Bhattacharya, and Ashutosh Modi. 2021.
\newblock \href {https://doi.org/10.18653/v1/2021.acl-long.313} {{ILDC} for {CJPE}: {I}ndian legal documents corpus for court judgment prediction and explanation}.
\newblock In \emph{Proceedings of the 59th Annual Meeting of the Association for Computational Linguistics and the 11th International Joint Conference on Natural Language Processing (Volume 1: Long Papers)}, pages 4046--4062, Online. Association for Computational Linguistics.

\bibitem[{Monroy et~al.(2009)Monroy, Calvo, and Gelbukh}]{monroy2009nlp}
Alfredo Monroy, Hiram Calvo, and Alexander Gelbukh. 2009.
\newblock \href {https://doi.org/10.1007/978-3-642-00382-0\_40} {{NLP} for shallow question answering of legal documents using graphs}.
\newblock In \emph{International Conference on Intelligent Text Processing and Computational Linguistics}, pages 498--508. Springer.

\bibitem[{Mukaka(2012)}]{mukaka2012guide}
Mavuto~M Mukaka. 2012.
\newblock A guide to appropriate use of correlation coefficient in medical research.
\newblock \emph{Malawi medical journal}, 24(3):69--71.

\bibitem[{Novikova et~al.(2017)Novikova, Dusek, Curry, and Rieser}]{DBLP:conf/emnlp/NovikovaDCR17}
Jekaterina Novikova, Ondrej Dusek, Amanda~Cercas Curry, and Verena Rieser. 2017.
\newblock \href {https://doi.org/10.18653/V1/D17-1238} {Why we need new evaluation metrics for {NLG}}.
\newblock In \emph{Proceedings of the 2017 Conference on Empirical Methods in Natural Language Processing, {EMNLP} 2017, Copenhagen, Denmark, September 9-11, 2017}, pages 2241--2252. Association for Computational Linguistics.

\bibitem[{Papineni et~al.(2002)Papineni, Roukos, Ward, and Zhu}]{papineni2002bleu}
Kishore Papineni, Salim Roukos, Todd Ward, and Wei-Jing Zhu. 2002.
\newblock \href {https://doi.org/10.3115/1073083.1073135} {Bleu: a method for automatic evaluation of machine translation}.
\newblock In \emph{Proceedings of the 40th annual meeting of the Association for Computational Linguistics}, pages 311--318.

\bibitem[{Sellam et~al.(2020)Sellam, Das, and Parikh}]{DBLP:conf/acl/SellamDP20}
Thibault Sellam, Dipanjan Das, and Ankur~P. Parikh. 2020.
\newblock \href {https://doi.org/10.18653/V1/2020.ACL-MAIN.704} {{BLEURT:} learning robust metrics for text generation}.
\newblock In \emph{Proceedings of the 58th Annual Meeting of the Association for Computational Linguistics, {ACL} 2020, Online, July 5-10, 2020}, pages 7881--7892. Association for Computational Linguistics.

\bibitem[{Wang et~al.(2018)Wang, Yang, Niu, Zhang, Zhang, and Niu}]{wang2018modeling}
Pengfei Wang, Ze~Yang, Shuzi Niu, Yongfeng Zhang, Lei Zhang, and ShaoZhang Niu. 2018.
\newblock \href {https://doi.org/10.1145/3209978.3210057} {Modeling dynamic pairwise attention for crime classification over legal articles}.
\newblock In \emph{the 41st international ACM SIGIR conference on research \& development in information retrieval}, pages 485--494.

\bibitem[{Wu et~al.(2020)Wu, Kuang, Zhang, Liu, Sun, Xiao, Zhuang, Si, and Wu}]{wu2020biased}
Yiquan Wu, Kun Kuang, Yating Zhang, Xiaozhong Liu, Changlong Sun, Jun Xiao, Yueting Zhuang, Luo Si, and Fei Wu. 2020.
\newblock \href {https://doi.org/10.18653/v1/2020.emnlp-main.56} {De-biased court's view generation with causality}.
\newblock In \emph{Proceedings of the 2020 Conference on Empirical Methods in Natural Language Processing, {EMNLP} 2020, Online, November 16-20, 2020}, pages 763--780. Association for Computational Linguistics.

\bibitem[{Wu et~al.(2023)Wu, Zhou, Liu, Lu, Liu, Zhang, Sun, Wu, and Kuang}]{wu2023precedent}
Yiquan Wu, Siying Zhou, Yifei Liu, Weiming Lu, Xiaozhong Liu, Yating Zhang, Changlong Sun, Fei Wu, and Kun Kuang. 2023.
\newblock \href {https://aclanthology.org/2023.emnlp-main.740} {Precedent-enhanced legal judgment prediction with {LLM} and domain-model collaboration}.
\newblock In \emph{Proceedings of the 2023 Conference on Empirical Methods in Natural Language Processing, {EMNLP} 2023, Singapore, December 6-10, 2023}, pages 12060--12075. Association for Computational Linguistics.

\bibitem[{Xiao et~al.(2021)Xiao, Hu, Liu, Tu, and Sun}]{xiao2021lawformer}
Chaojun Xiao, Xueyu Hu, Zhiyuan Liu, Cunchao Tu, and Maosong Sun. 2021.
\newblock \href {https://doi.org/10.1016/J.AIOPEN.2021.06.003} {Lawformer: A pre-trained language model for chinese legal long documents}.
\newblock \emph{AI Open}, 2:79--84.

\bibitem[{Xiong(2021)}]{Xiong2021Relationship}
Qiuhong Xiong. 2021.
\newblock \href {https://doi.org/10.19430/j.cnki.3891.2021.01.002} {On relationship of public prosecution and private prosecution}.
\newblock \emph{Criminal Science}.

\bibitem[{Xu et~al.(2020)Xu, Wang, Chen, Pan, Wang, and Zhao}]{xu2020distinguish}
Nuo Xu, Pinghui Wang, Long Chen, Li~Pan, Xiaoyan Wang, and Junzhou Zhao. 2020.
\newblock \href {https://doi.org/10.18653/V1/2020.ACL-MAIN.280} {Distinguish confusing law articles for legal judgment prediction}.
\newblock In \emph{Proceedings of the 58th Annual Meeting of the Association for Computational Linguistics, {ACL} 2020, Online, July 5-10, 2020}, pages 3086--3095. Association for Computational Linguistics.

\bibitem[{Yang et~al.(2024)Yang, Yang, Zhang, Hui, Zheng, Yu, Li, Liu, Huang, Wei, Lin, Yang, Tu, Zhang, Yang, Yang, Zhou, Lin, Dang, Lu, Bao, Yang, Yu, Li, Xue, Zhang, Zhu, Men, Lin, Li, Xia, Ren, Ren, Fan, Su, Zhang, Wan, Liu, Cui, Zhang, and Qiu}]{qwen2.5}
An~Yang, Baosong Yang, Beichen Zhang, Binyuan Hui, Bo~Zheng, Bowen Yu, Chengyuan Li, Dayiheng Liu, Fei Huang, Haoran Wei, Huan Lin, Jian Yang, Jianhong Tu, Jianwei Zhang, Jianxin Yang, Jiaxi Yang, Jingren Zhou, Junyang Lin, Kai Dang, and 22 others. 2024.
\newblock \href {https://doi.org/10.48550/ARXIV.2412.15115} {Qwen2.5 technical report}.
\newblock \emph{CoRR}, abs/2412.15115.

\bibitem[{Zhang et~al.(2020)Zhang, Kishore, Wu, Weinberger, and Artzi}]{zhang2019bertscore}
Tianyi Zhang, Varsha Kishore, Felix Wu, Kilian~Q. Weinberger, and Yoav Artzi. 2020.
\newblock \href {https://openreview.net/forum?id=SkeHuCVFDr} {Bertscore: Evaluating text generation with {BERT}}.
\newblock In \emph{8th International Conference on Learning Representations, {ICLR} 2020, Addis Ababa, Ethiopia, April 26-30, 2020}. OpenReview.net.

\bibitem[{Zhang et~al.(2023)Zhang, Hu, Zhang, Zhang, Wang, Qu, and Xu}]{zhang-etal-2023-fedlegal}
Zhuo Zhang, Xiangjing Hu, Jingyuan Zhang, Yating Zhang, Hui Wang, Lizhen Qu, and Zenglin Xu. 2023.
\newblock \href {https://doi.org/10.18653/v1/2023.acl-long.193} {{FEDLEGAL}: The first real-world federated learning benchmark for legal {NLP}}.
\newblock In \emph{Proceedings of the 61st Annual Meeting of the Association for Computational Linguistics (Volume 1: Long Papers)}, pages 3492--3507, Toronto, Canada. Association for Computational Linguistics.

\bibitem[{Zhao et~al.(2021)Zhao, Yue, An, Liu, Zhang, He, Chen, Yuan, and Liu}]{zhao2021legal}
Lili Zhao, Linan Yue, Yanqing An, Ye~Liu, Kai Zhang, Weidong He, Yanmin Chen, Senchao Yuan, and Qi~Liu. 2021.
\newblock \href {https://doi.org/10.1007/978-3-030-93046-2\_60} {Legal judgment prediction with multiple perspectives on civil cases}.
\newblock In \emph{Artificial Intelligence: First CAAI International Conference, CICAI 2021, Hangzhou, China, June 5--6, 2021, Proceedings, Part I 1}, pages 712--723. Springer.

\bibitem[{Zhao et~al.(2022)Zhao, Yue, An, Zhang, Yu, Liu, and Chen}]{zhao2022cpee}
Lili Zhao, Linan Yue, Yanqing An, Yuren Zhang, Jun Yu, Qi~Liu, and Enhong Chen. 2022.
\newblock \href {https://doi.org/10.1145/3511808.3557273} {{CPEE:} civil case judgment prediction centering on the trial mode of essential elements}.
\newblock In \emph{Proceedings of the 31st ACM International Conference on Information \& Knowledge Management}, pages 2691--2700.

\bibitem[{Zhong et~al.(2018)Zhong, Guo, Tu, Xiao, Liu, and Sun}]{zhong2018legal}
Haoxi Zhong, Zhipeng Guo, Cunchao Tu, Chaojun Xiao, Zhiyuan Liu, and Maosong Sun. 2018.
\newblock \href {https://doi.org/10.18653/V1/D18-1390} {Legal judgment prediction via topological learning}.
\newblock In \emph{Proceedings of the 2018 conference on empirical methods in natural language processing}, pages 3540--3549.

\bibitem[{Zhong et~al.(2020{\natexlab{a}})Zhong, Xiao, Tu, Zhang, Liu, and Sun}]{zhong2020does}
Haoxi Zhong, Chaojun Xiao, Cunchao Tu, Tianyang Zhang, Zhiyuan Liu, and Maosong Sun. 2020{\natexlab{a}}.
\newblock \href {https://doi.org/10.18653/V1/2020.ACL-MAIN.466} {How does {NLP} benefit legal system: {A} summary of legal artificial intelligence}.
\newblock In \emph{Proceedings of the 58th Annual Meeting of the Association for Computational Linguistics, {ACL} 2020, Online, July 5-10, 2020}, pages 5218--5230. Association for Computational Linguistics.

\bibitem[{Zhong et~al.(2020{\natexlab{b}})Zhong, Xiao, Tu, Zhang, Liu, and Sun}]{zhong2020jec}
Haoxi Zhong, Chaojun Xiao, Cunchao Tu, Tianyang Zhang, Zhiyuan Liu, and Maosong Sun. 2020{\natexlab{b}}.
\newblock \href {https://doi.org/10.1609/AAAI.V34I05.6519} {{JEC-QA:} {A} legal-domain question answering dataset}.
\newblock In \emph{The Thirty-Fourth {AAAI} Conference on Artificial Intelligence, {AAAI} 2020, The Thirty-Second Innovative Applications of Artificial Intelligence Conference, {IAAI} 2020, The Tenth {AAAI} Symposium on Educational Advances in Artificial Intelligence, {EAAI} 2020, New York, NY, USA, February 7-12, 2020}, pages 9701--9708. {AAAI} Press.

\end{thebibliography}

\appendix


\begin{table*}[ht]
\small
\centering
\begin{tabular}{lll}
\toprule[2pt]
\textbf{Model}                      & \textbf{Version} & \textbf{Context Window}  \\ \midrule
                        \midrule
GPT-4o \citep{hurst2024gpt}                 &  gpt-4o-2024-08-06          &  128,000 tokens       \\
LLaMA3.1 \citep{grattafiori2024llama}                  &  llama-3.1-8b-instruct          &  131,072 tokens     \\
Claude3.5 \citep{anthropic2024claude35addendum}                  &  Claude-3-5-sonnet-20241022          &  200,000 tokens                \\
Qwen2.5 \citep{qwen2.5}                         &     Qwen2.5-7b-instruct &  128,000 tokens     \\
DeepSeek-R1 \citep{guo2025deepseek}                        &     deepseek-reasoner            &  64,000 tokens      \\
Farui \citep{alibaba2024farui}             &   farui-plus    &  12,000 tokens
  \\\bottomrule
\end{tabular}
\caption{Details of used LLMs.}
\label{tab:LLMs}
\end{table*}

\section{Details of LLMs Used in This Paper}
\label{app:llm}

As shown in Table \ref{tab:LLMs}, we employ six LLMs: GPT-4o \citep{hurst2024gpt}, LLaMA3.1 \citep{grattafiori2024llama}, Claude3.5 \citep{anthropic2024claude35addendum}, Qwen2.5 \citep{qwen2.5}, DeepSeek-R1 \citep{guo2025deepseek} and Farui \citep{alibaba2024farui}.
The version of GPT-4o is gpt-4o-2024-08-06, the snapshot of gpt-4o from August 6th 2024. As for LLaMA3.1, we use the llama-3.1-8b-instruct  version.The version of Claude3.5 that we use is Claude-3-5-sonnet-20241022, and the model was last updated in April 2024. We use Qwen2.5-7b-instruct as the version of Qwen, which is an instruction tuning model that performs well on reasoning, coding and mathematics tasks.The version of DeepSeek-R1 that we utilize is deepseek-reasoner, and Farui's version is farui-plus.  GPT-4o has a context window of 128,000 tokens. LLaMA3.1 (specifically the llama-3.1-8b-instruct version) has a context window of 131,072 tokens, surpassing that of Claude3.5 (200,000 tokens) and Qwen2.5 (128,000 tokens). DeepSeek-R1 features a context window of 64,000 tokens, and Farui has the smallest context window among them with 12,000 tokens. Each model's version and characteristics are tailored to meet different application needs and scenarios.  

\section{API Access URLs}
\label{app:urls}

\vspace{0.5em}
\begin{itemize}
    \setlength\itemsep{0.5em} 
    \item \hangindent=2em \textbf{GPT-4o (OpenAI):} \texttt{\url{https://platform.openai.com/docs/models/gpt-4o}}
    \item \hangindent=2em \textbf{Claude 3.5 (Anthropic):} \texttt{\url{https://docs.anthropic.com/en/docs/about-claude/models/overview}}
    \item \hangindent=2em \textbf{DeepSeek-R1 (Deepseek):} \texttt{\url{https://api-docs.deepseek.com/}}
    \item \hangindent=2em \textbf{Farui (Aliyun):} \texttt{\url{https://tongyi.aliyun.com/farui/guide/api_description_doc}}
    \item \hangindent=2em \textbf{LLaMA3.1 (Hugging Face):} \texttt{\url{https://huggingface.co/meta-llama/Llama-3.1-8B-Instruct}}
    \item \hangindent=2em \textbf{Qwen2.5 (Hugging Face):} \texttt{\url{https://huggingface.co/Qwen/Qwen2.5-7B-Instruct}}
\end{itemize}

\section{GPT-4o Scoring Prompts}
\label{sec:GPT4ScoringPrompts}
The detailed prompts for two metrics are as follows.
\paragraph{For factuality.}\textit{``Factuality in claims requires that statements of fact be truthful, precise, and based on objectively existing circumstances. Please rate from 1 to 5 based on factual accuracy. Deduct points for errors by the plaintiff or defendant, discrepancies between factual descriptions and real scenarios, and inconsistencies between claims and actual events, whether adding or omitting details.''}
\paragraph{For clarity.}\textit{``Clarity refers to statements that should be clear, concise, and unambiguous, avoiding vagueness, ambiguity, and unnecessary redundancy. Please rate from 1 to 5 based on clarity. Deduct points for repetitive or redundant reasoning sections and claims. Excessive additional information will result in a deduction of points.''}

\section{Details about Human Annotation}

\subsection{Full Annotation Guidelines Provided to Annotators}
\label{sec:annotation-guidelines}

To ensure consistency and fairness in human evaluation, we provided all annotators with a standardized set of instructions. Below, we report the full text of the annotation guidelines given to participants, in accordance with ethical best practices. This includes the task description, scoring rubric, and any disclaimers or risk notices if applicable.

\vspace{0.5em}
\noindent\textbf{Annotation Task Description} \\
Annotators were instructed to evaluate each model-generated legal claim based on two criteria: \textit{Factuality} and \textit{Clarity}. Definitions of each are given below.

\begin{itemize}
    \item \textbf{Factuality (1–5):} Factuality in claims requires that statements of fact be truthful, precise, and based on objectively existing circumstances. Please rate from 1 to 5 based on factual accuracy. Deduct points for errors by the plaintiff or defendant, discrepancies between factual descriptions and real scenarios, and inconsistencies between claims and actual events, whether adding or omitting details.
    \item \textbf{Clarity (1–5):} Clarity refers to statements that should be clear, concise, and unambiguous, avoiding vagueness, ambiguity, and unnecessary redundancy. Please rate from 1 to 5 based on clarity. Deduct points for repetitive or redundant reasoning sections and claims. Excessive additional information will result in a deduction of points.
\end{itemize}

\noindent\textbf{Scoring Scale} \\
Annotators were instructed to assign integer scores from 1 (very poor) to 5 (excellent), with intermediate scores indicating partial satisfaction of the criterion.

\noindent\textbf{Disclaimer} \\
Annotators were informed that the task involved reading real legal case descriptions and that no personally identifiable information (PII) was included. They were told they could opt out at any time. No personal risks were identified for participants in this task.

\begin{figure*}[ht]
    \centering
    \includegraphics[width=1\textwidth]{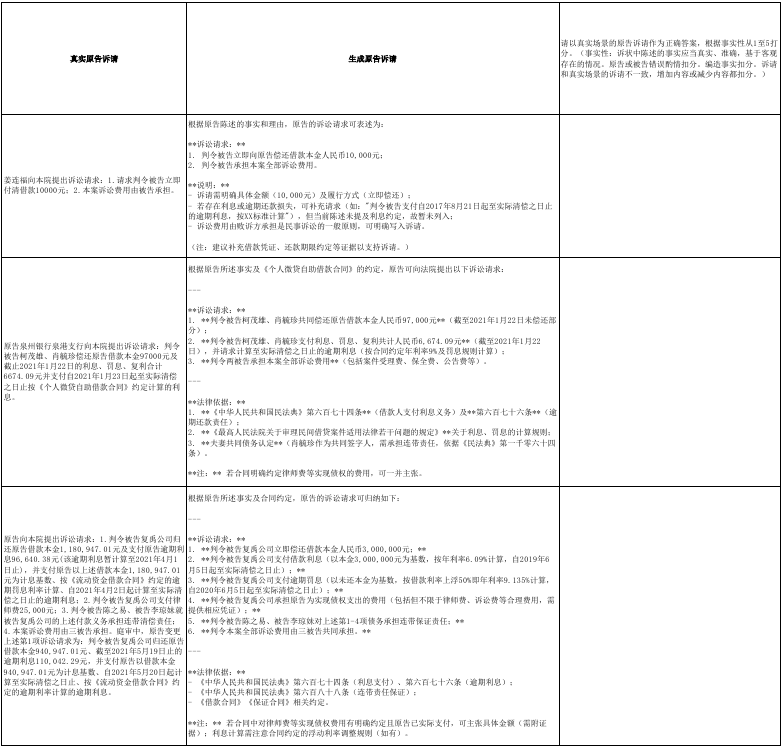}
    \caption{Interface used for human annotation.}
    \label{fig:annotation-ui}
\end{figure*}

Figure~\ref{fig:annotation-ui} shows the interface provided to annotators for scoring legal claims generated by the model.

\subsection{Evaluation Robustness and Metric Validity}
\label{sec:MetaEvaluation}

Meta-evaluation of the proposed metrics on SummEval \citep{DBLP:journals/tacl/FabbriKMXSR21} includes the Spearman correlation coefficient and Pearson correlation coefficient. The Spearman correlation coefficient and Pearson correlation coefficient between human and GPT-4o scores are respectively 0.5197 and 0.5248, which means a moderate correlation \citep{mukaka2012guide}.

\subsection{Annotation Protocol and Inter-Annotator Agreement}
\label{sec:annotation-protocol}
The inter-annotator agreement (IAA) between annotators and the correlation between human and GPT-4o scores are detailed as follows. The Fleiss’ Kappa among 3 annotators is 0.6823, indicating substantial agreement according to Landis and Koch’s interpretation scale \citep{landis1977measurement}, which validates the human evaluation process. The Spearman correlation coefficient between human and GPT-4o scores is 0.5197, and the Pearson correlation coefficient is 0.5248, both of which indicate a moderate correlation \citep{mukaka2012guide}.

\subsection{Metric Definitions and Scoring Procedure}
\label{sec:HumanEvaluation}

We provide the formal definitions of the evaluation metrics used in Section~\ref{sec:human-eval}, including the normalization process and the calculation of mean absolute error (MAE) and consistency.

\vspace{0.5em}
\noindent\textbf{Normalization.} Given raw human and model scores \( J_i, P_i \in [s_{\min}, s_{\max}] \), we normalize them to the \([0,1]\) interval as follows:
\[
\tilde{J}_i = \frac{J_i - s_{\min}}{s_{\max} - s_{\min}}, \quad \tilde{P}_i = \frac{P_i - s_{\min}}{s_{\max} - s_{\min}}
\]

\vspace{0.5em}
\noindent\textbf{Metric Computation.} Let \( n \) be the number of evaluated samples. The metrics are computed as:
\[
\text{MAE} = \frac{1}{n} \sum_{i=1}^{n} \left| \tilde{J}_i - \tilde{P}_i \right|
\]
\[
\text{Consistency} = \frac{1}{n} \sum_{i=1}^{n} \mathbf{1} \left( \left| \tilde{J}_i - \tilde{P}_i \right| \leq \frac{\delta}{s_{\max} - s_{\min}} \right)
\]

\vspace{0.5em}
\noindent In our experiments, we use a 5-point rating scale where \( s_{\min} = 1 \), \( s_{\max} = 5 \), and the tolerance threshold \( \delta = 1 \). Thus, consistency corresponds to the proportion of samples where the absolute difference between model and human scores does not exceed 1.

\section{Case Examples for Error Analysis}
\label{sec:cases}

We provide the full input and model-generated claims for the representative cases referenced in Section~\ref{sec:error-analysis}. For each case, we list the input facts (prompt''), the ground truth legal claim written by human experts (ground truth’’), and the responses generated by each model. The selected examples include \textit{CaseID 1, 2, 31, 93, 98, 110, and 116}, covering diverse types of factual patterns and model errors.

To facilitate close comparison, we highlight key elements in color:
\textcolor{blue}{\underline{\textbf{Blue text}}} indicates critical factual expressions within the prompt.
\textcolor{darkgreen}{\underline{\textbf{Green text}}} marks the key elements in the ground truth claim that correspond to the legal intent.
\textcolor{red}{\underline{\textbf{Red text}}} is used to highlight incorrect or hallucinated content within the model outputs.
\textcolor{mypurple}{\underline{\textbf{Purple text}}} highlights correct legal reasoning or factual reproduction by the models.

These annotated examples serve as the basis for our multi-dimensional error analysis in the main text, enabling a fine-grained understanding of how different models succeed or fail in legal claim generation.
\begin{figure*}[ht]
    \centering
    \includegraphics[height=0.9\textheight, keepaspectratio]{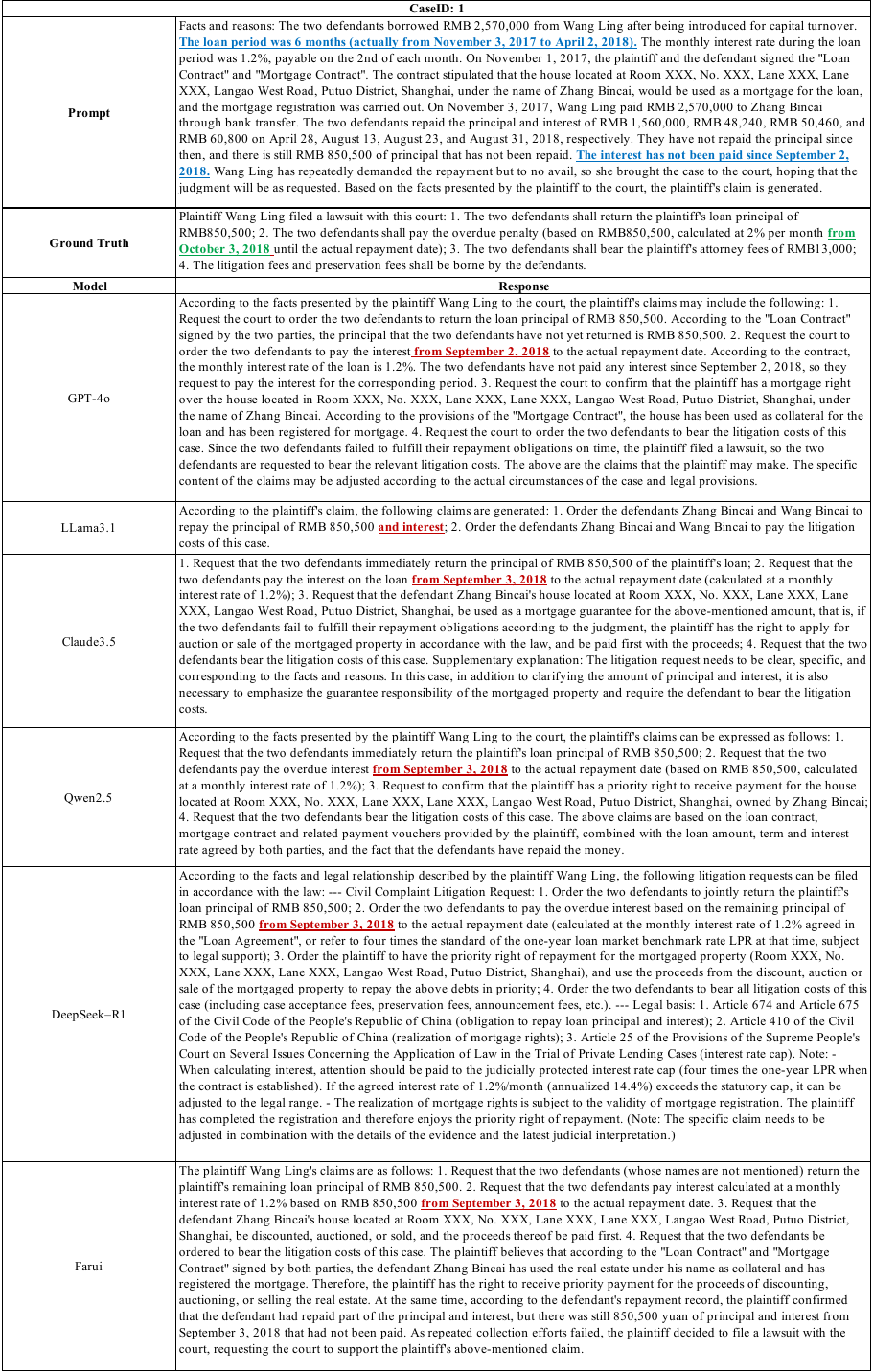}
    \caption{CaseId~1.}
\end{figure*}

\begin{figure*}[ht]
    \centering
    \includegraphics[height=0.9\textheight, keepaspectratio]{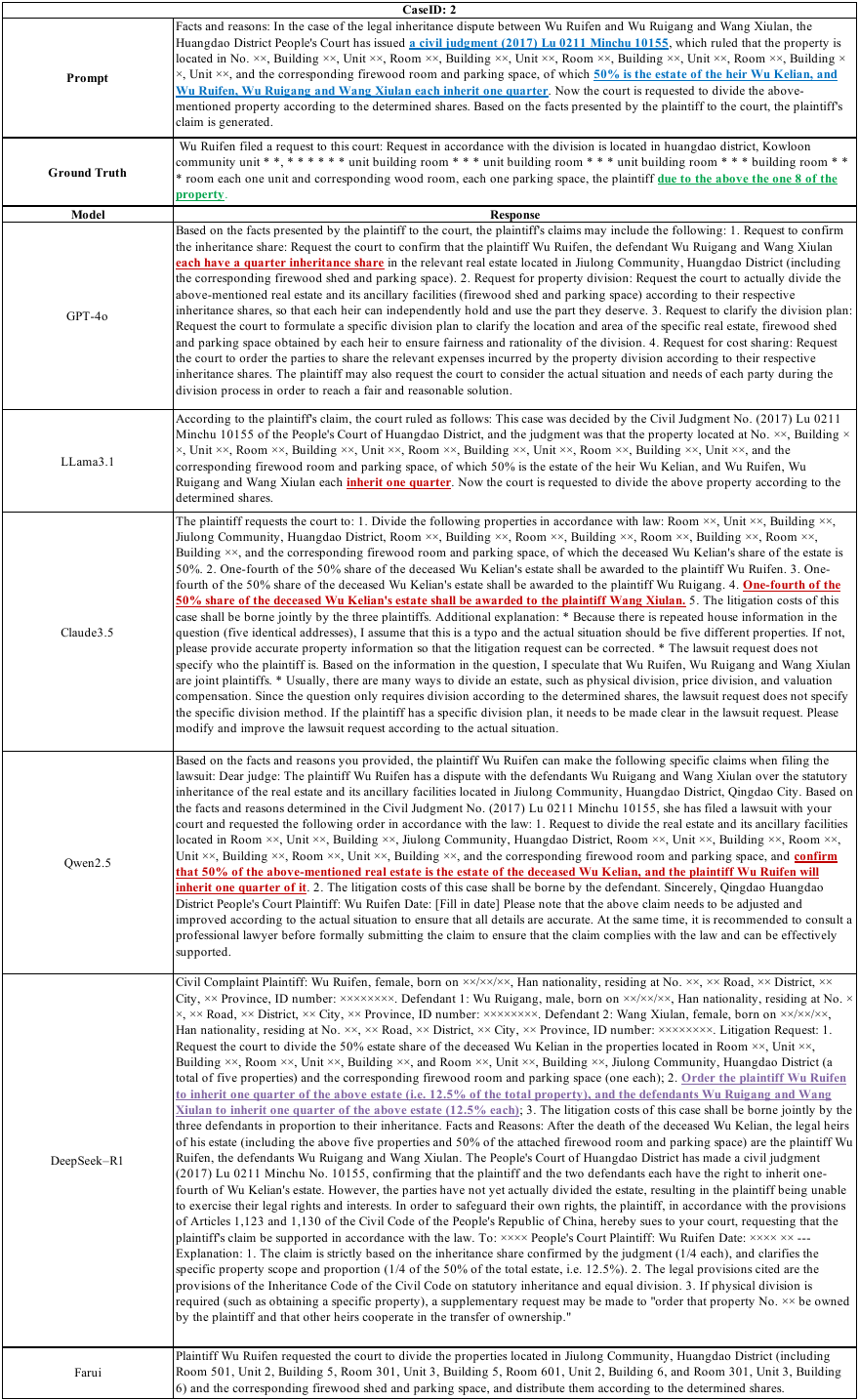}
    \caption{CaseId~2.}
\end{figure*}

\begin{figure*}[ht]
    \centering
    \includegraphics[height=0.9\textheight, keepaspectratio]{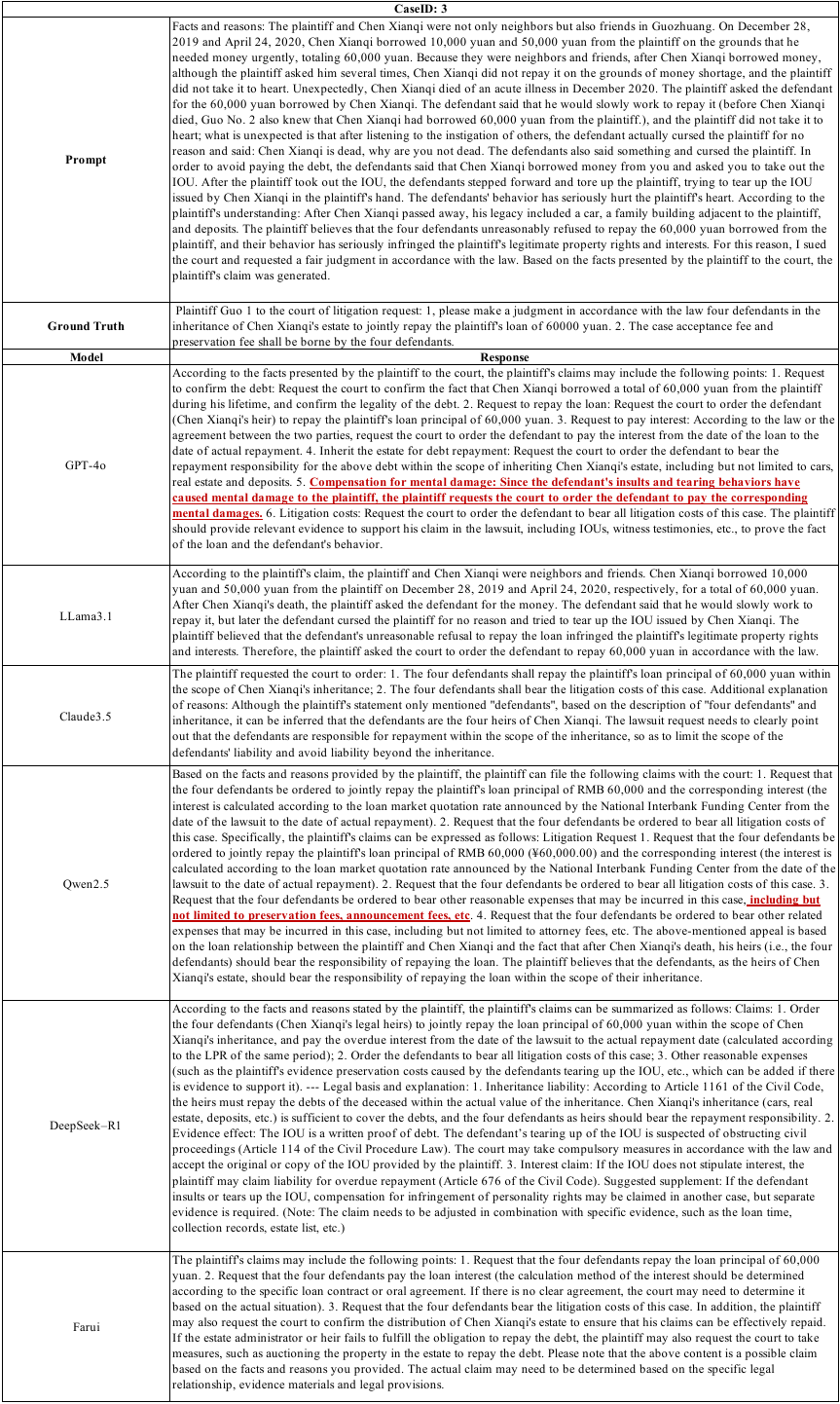}
    \caption{CaseId~3.}
\end{figure*}

\begin{figure*}[ht]
    \centering
    \includegraphics[height=0.9\textheight, keepaspectratio]{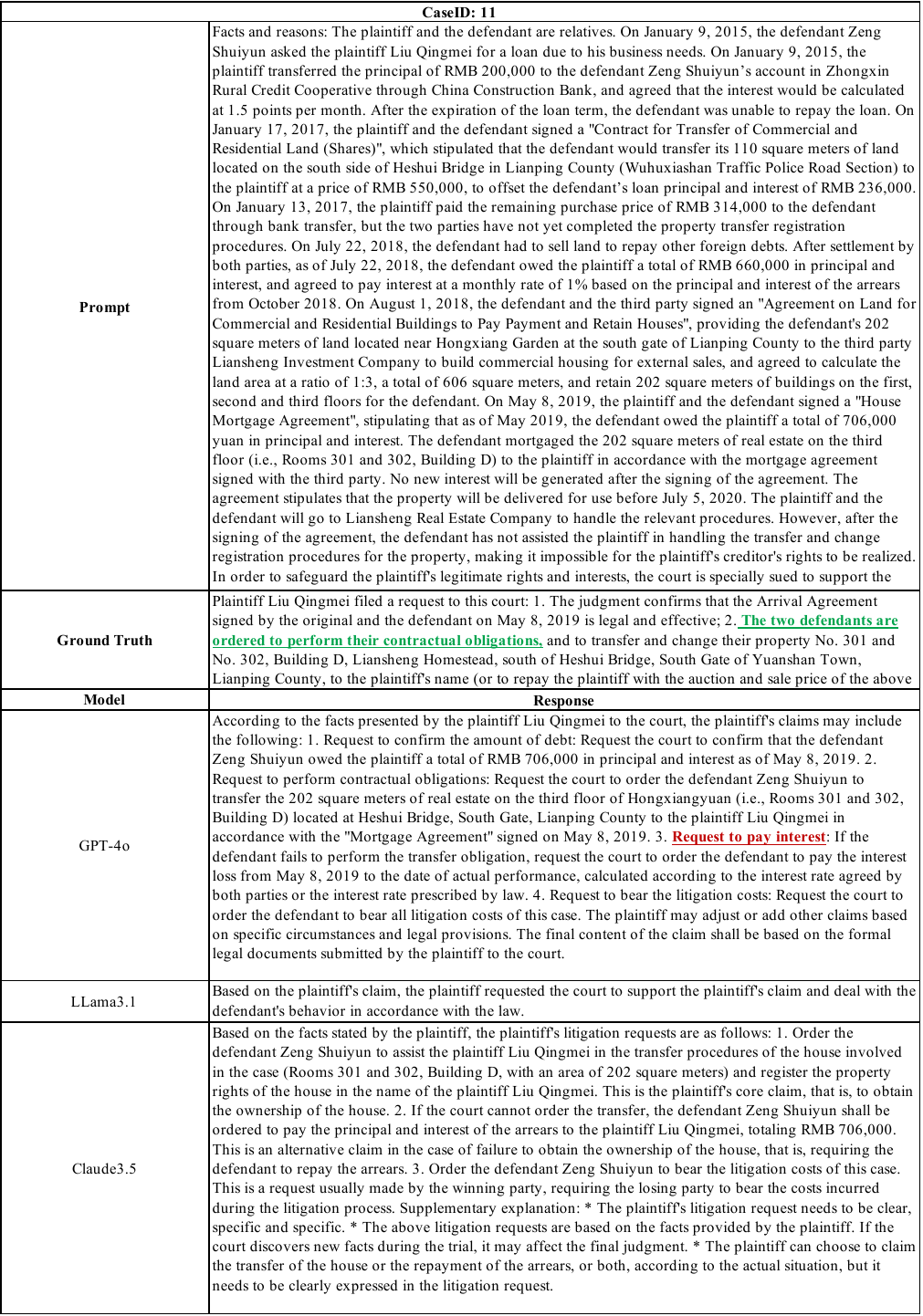}
    \caption{CaseId~11-part1.}
\end{figure*}

\begin{figure*}[ht]
    \centering
    \includegraphics[height=0.9\textheight, keepaspectratio]{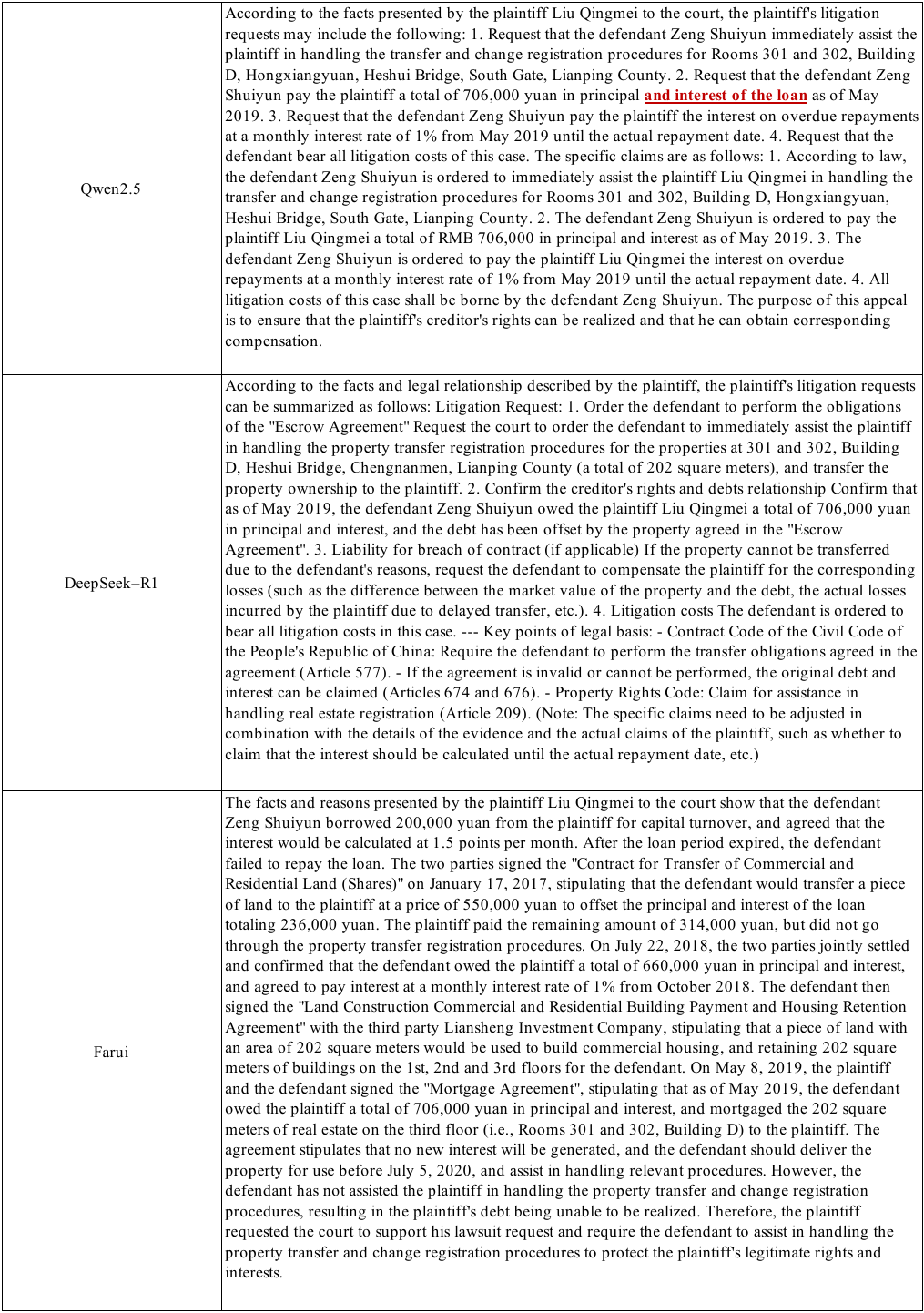}
    \caption{CaseId~11-part2.}
\end{figure*}

\begin{figure*}[ht]
    \centering
    \includegraphics[height=0.9\textheight, keepaspectratio]{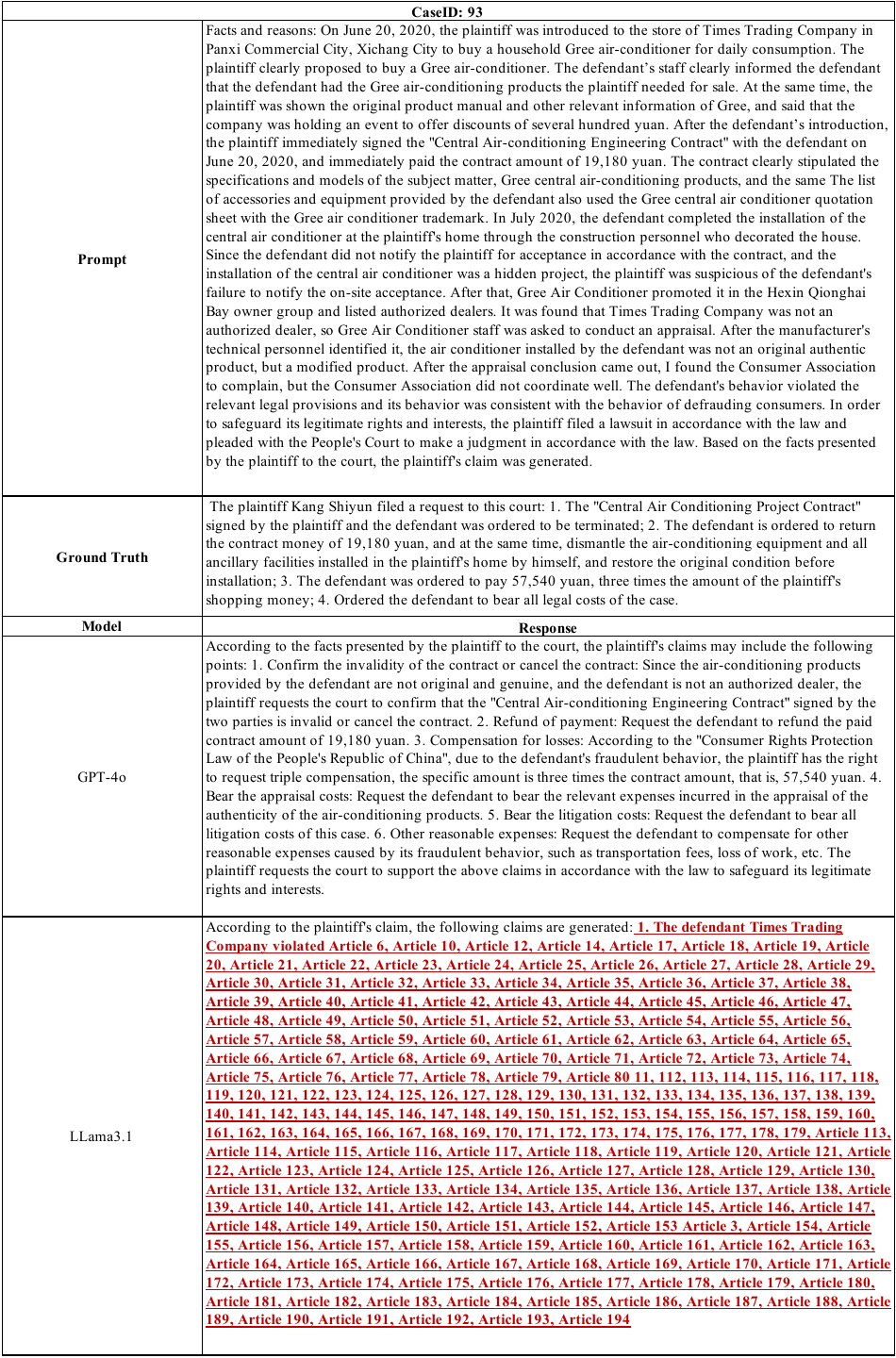}
    \caption{CaseId~93-part1.}
\end{figure*}

\begin{figure*}[ht]
    \centering
    \includegraphics[height=0.9\textheight, keepaspectratio]{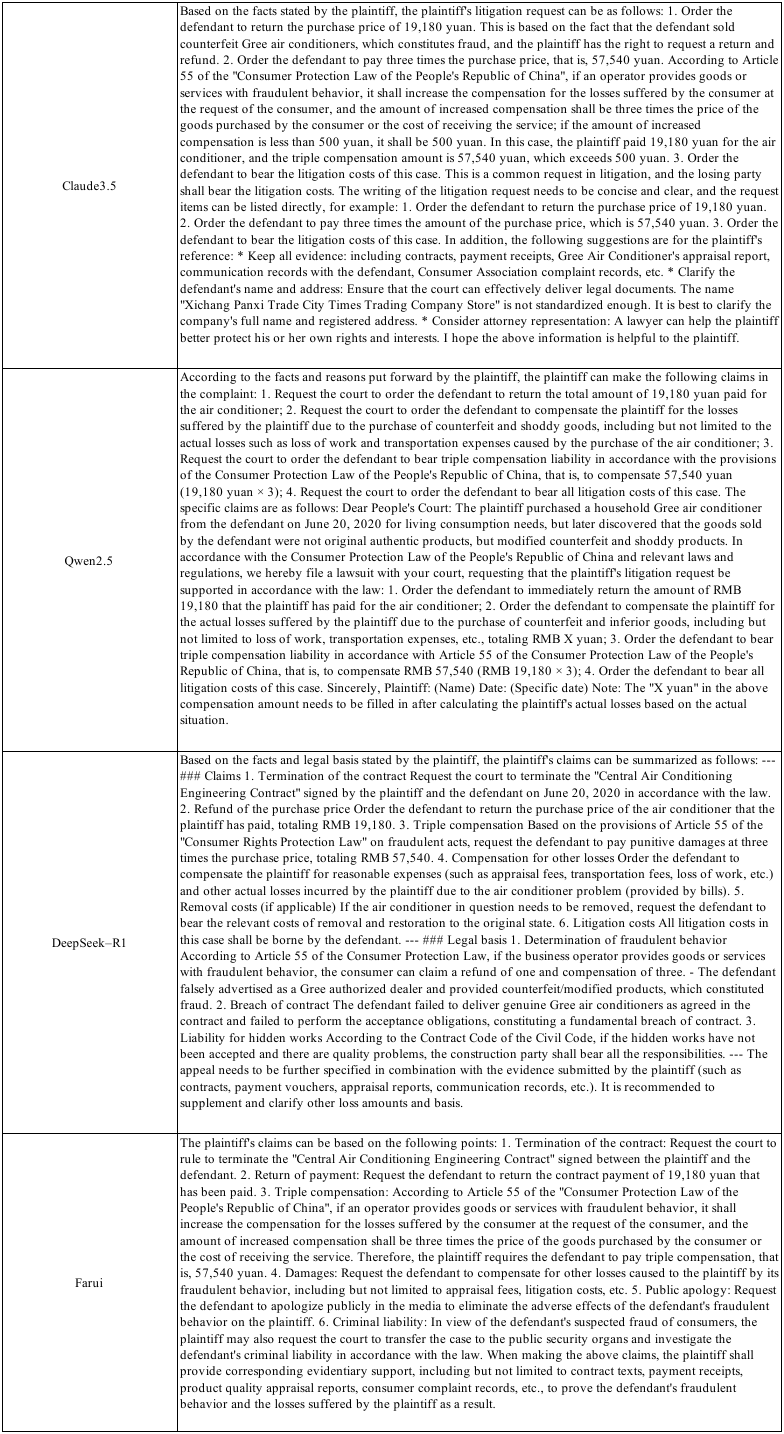}
    \caption{CaseId~93-part2.}
\end{figure*}

\begin{figure*}[ht]
    \centering
    \includegraphics[height=0.9\textheight, keepaspectratio]{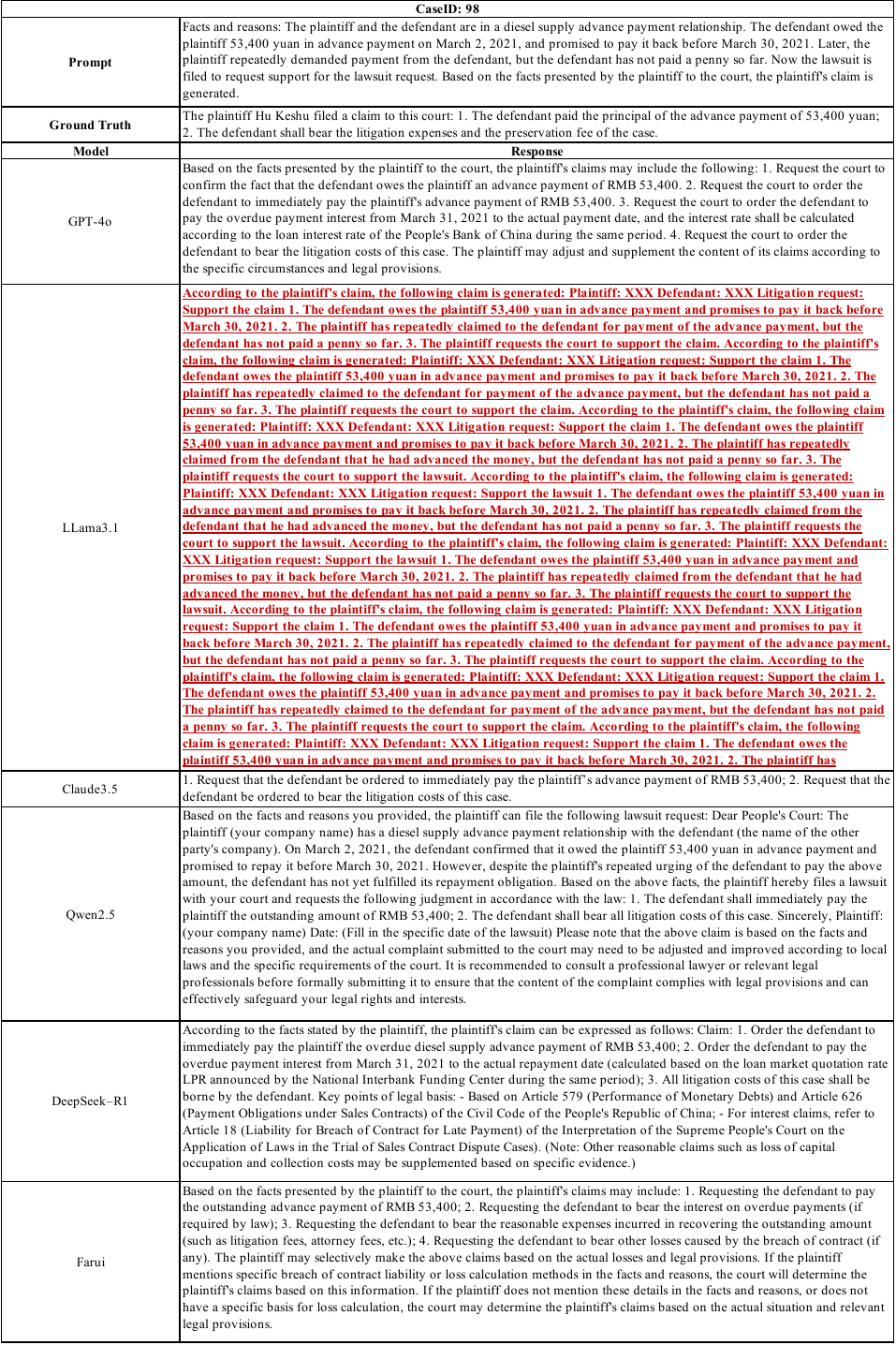}
    \caption{CaseId~98.}
\end{figure*}

\begin{figure*}[ht]
    \centering
    \includegraphics[height=0.9\textheight, keepaspectratio]{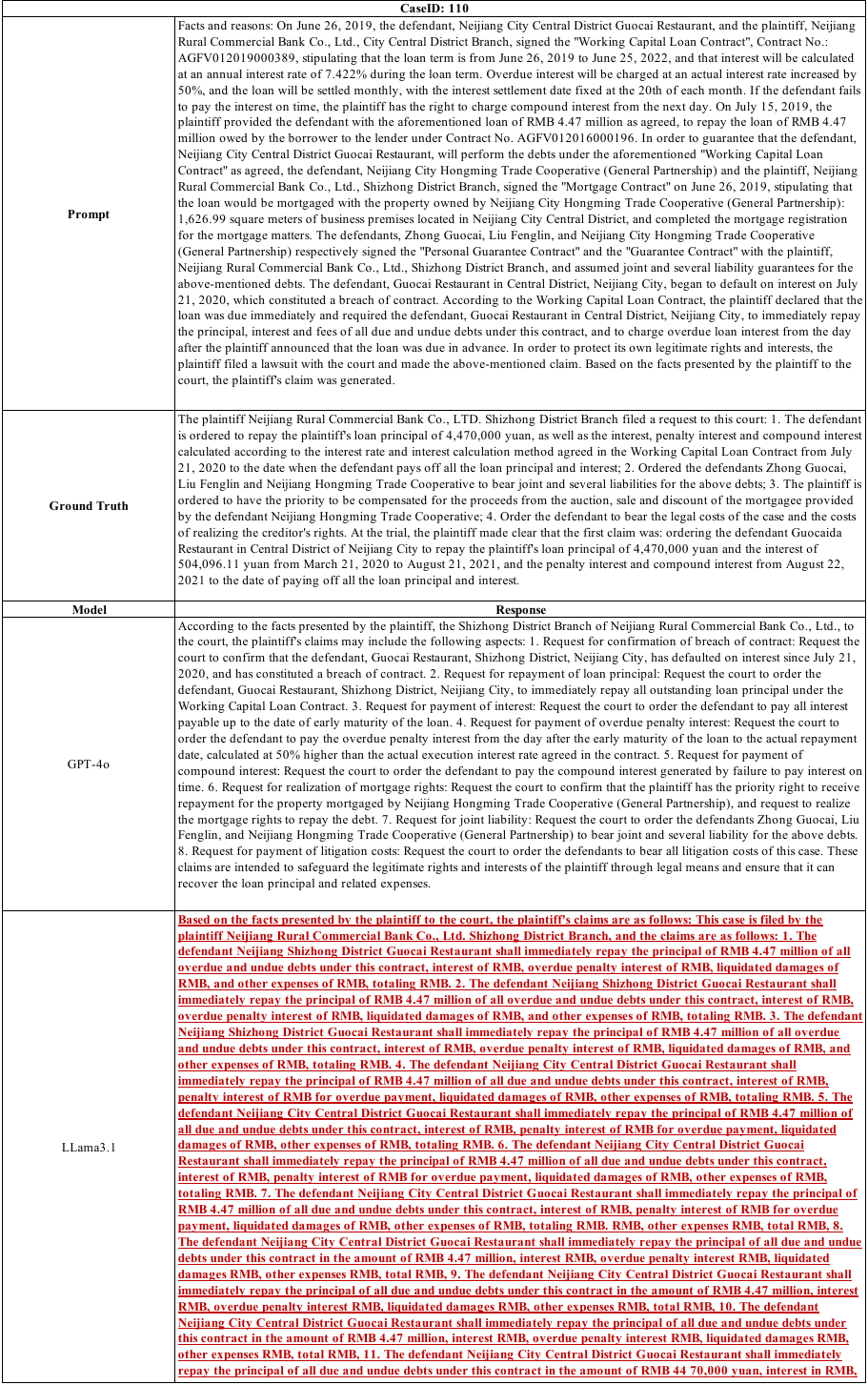}
    \caption{CaseId~110-part1.}
\end{figure*}

\begin{figure*}[ht]
    \centering
    \includegraphics[height=0.9\textheight, keepaspectratio]{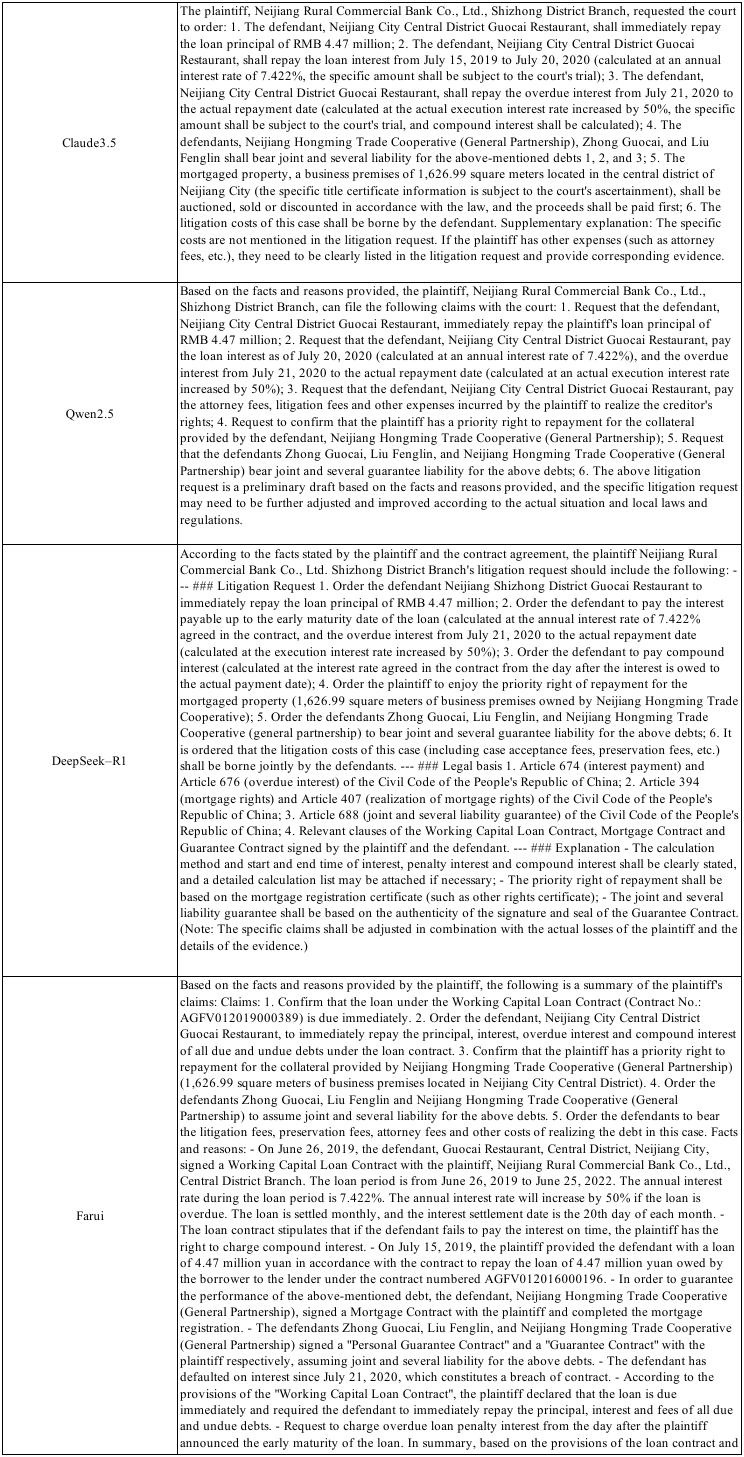}
    \caption{CaseId~110-part2.}
\end{figure*}

\begin{figure*}[ht]
    \centering
    \includegraphics[height=0.9\textheight, keepaspectratio]{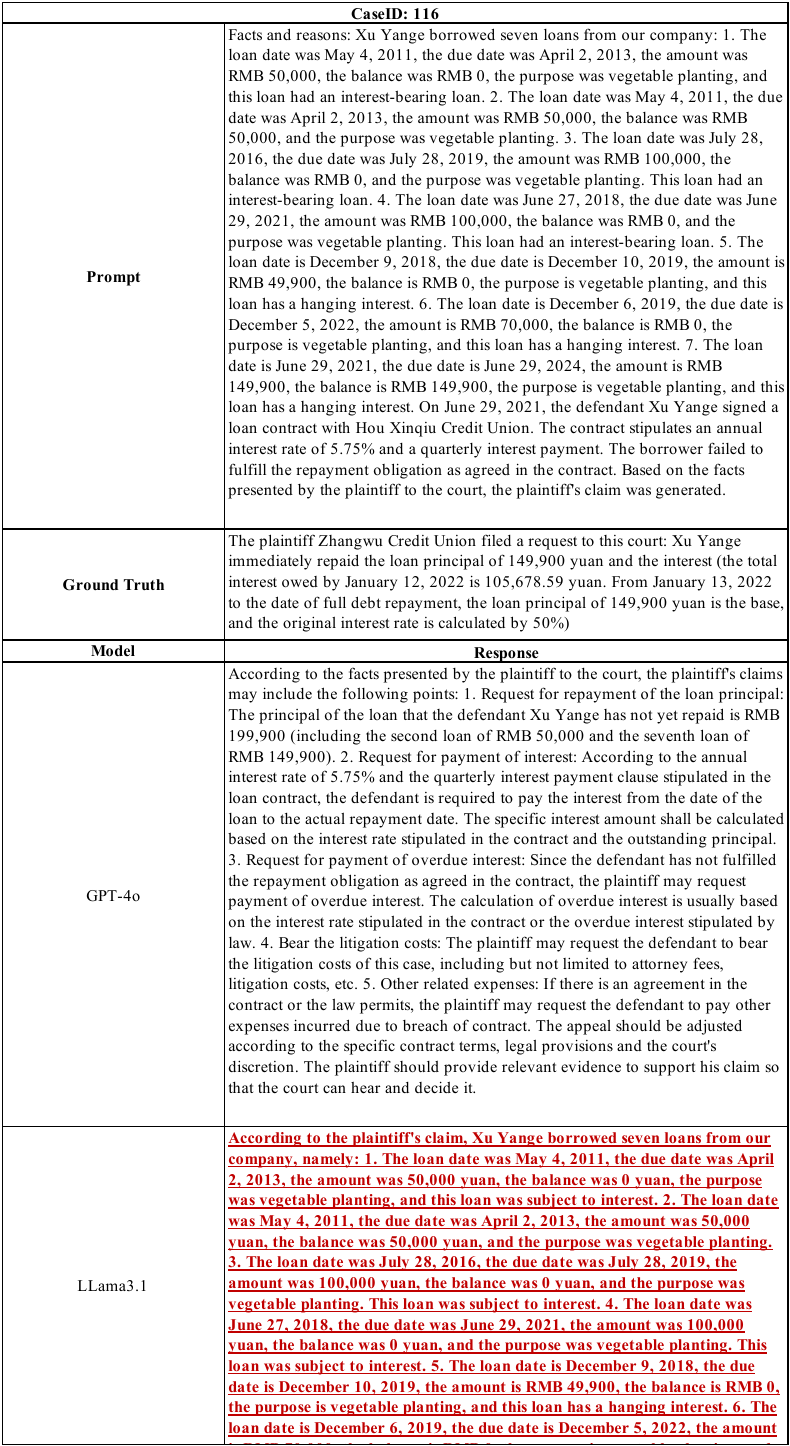}
    \caption{CaseId~116-part1.}
\end{figure*}

\begin{figure*}[ht]
    \centering
    \includegraphics[height=0.9\textheight, keepaspectratio]{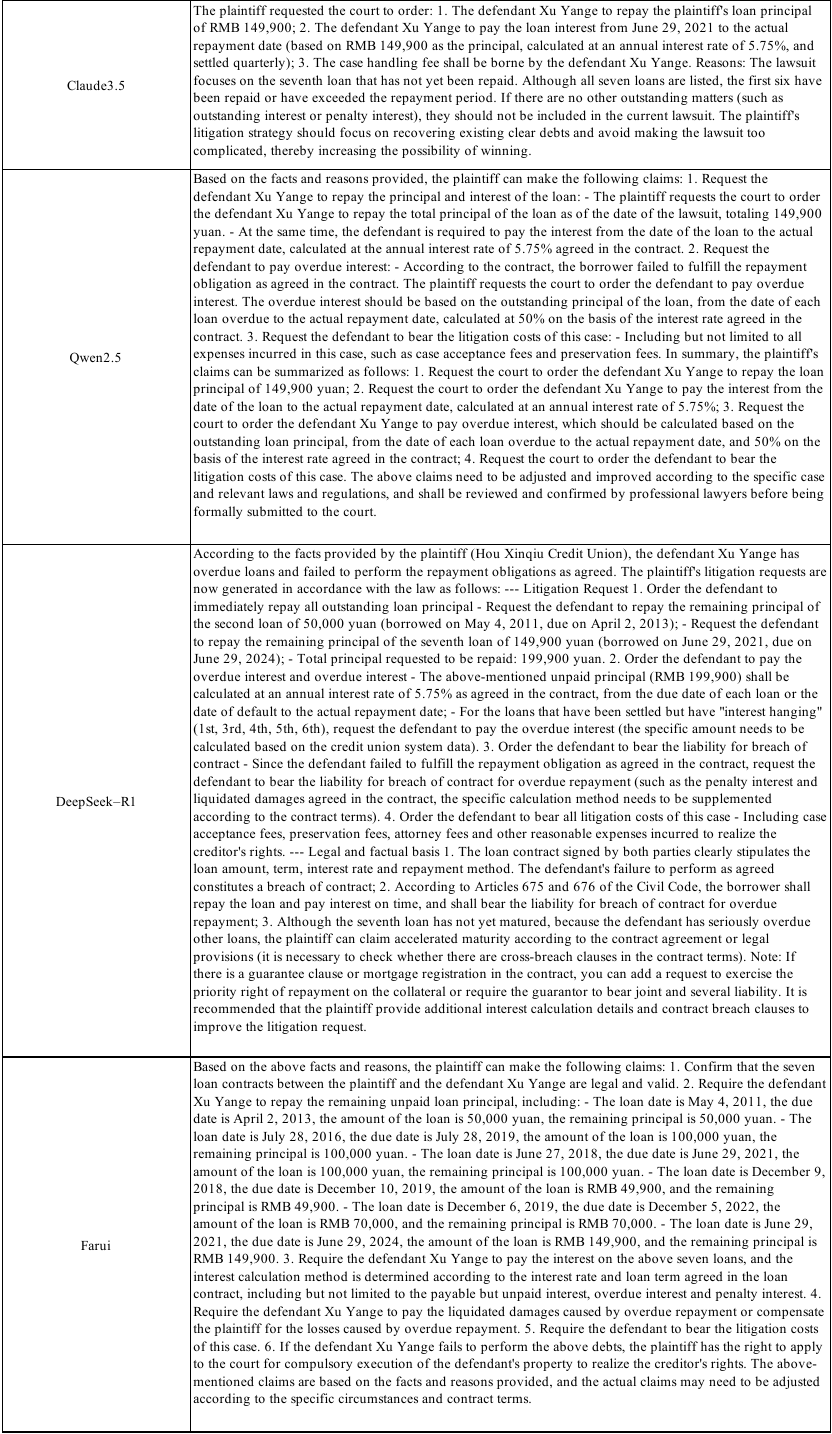}
    \caption{CaseId~116-part2.}
\end{figure*}

\end{document}